\documentclass[11pt]{article}
\usepackage{enumitem}

\usepackage{booktabs}
\usepackage{makecell}
\usepackage{xspace}
\usepackage{array}

\newcolumntype{L}[1]{>{\raggedright\arraybackslash}p{#1}}
\newcolumntype{C}[1]{>{\centering\arraybackslash}p{#1}}

\usepackage{xurl}
\usepackage{hyperref}
\usepackage[preprint]{acl}

\usepackage{times}
\usepackage{latexsym}

\usepackage{amsmath}
\usepackage{amssymb}

\usepackage[T1]{fontenc}
\usepackage[utf8]{inputenc}

\usepackage{microtype}

\usepackage{inconsolata}

\usepackage{graphicx}

\newcommand{\QuoteUnquote}[1]{``#1''}
\newcommand{\NumDatasets}{46}
\newcommand{\NumAlgos}{15}
\newcommand{\NumAlgoFamilies}{6}
\newcommand{\NumDatasetFamilies}{16}
\newcommand{\TIDDatasetCount}{4}
\newcommand{\RMISS}{R_{\mathrm{MISS}}}
\newcommand{\MultitaskEpochs}{30}

\makeatletter
\newcommand{\ComputeAcknowledgment}{%
  \ifacl@anonymize
    institutionally managed high-performance computing resources%
  \else
    the Minnesota Supercomputing Institute (MSI) at the University of Minnesota%
  \fi
}

\newcommand{\MSIAcknowledgment}{%
  \ifacl@anonymize
  \else
    The authors acknowledge the Minnesota Supercomputing Institute (MSI)
    at the University of Minnesota for providing resources that contributed
    to the research results reported within this paper.%
  \fi
}
\makeatother

\title{Who Do Perspectivist Models Predict Well? Annotator Agreement, Sparsity, and Performance}

\author{
    {\bf London Lowmanstone IV} \quad
    {\bf Vaagishwaran Sivakumar} \quad
    {\bf Anwesha Samaddar} \\
    {\bf Akul Mundada} \quad
    {\bf Harnoor Kaur} \quad
    {\bf Maria Gini} \\
    University of Minnesota, Twin Cities
}

\begin{document}

\maketitle

\begin{abstract}
Successful completion of perspectivist natural language processing (NLP) annotation tasks requires algorithms to be aligned across diverse perspectives in order to predict how different humans will rate text. This work studies how the performance of such algorithms varies for annotators who behave differently than others, either by providing annotations that diverge from others' responses or by providing fewer annotations altogether. To aid with characterizing model performance, we curate the largest collection to date of perspectivist NLP datasets with ordinal annotations. After evaluating multiple classes of perspectivist algorithms across each of the datasets, we model the relationship between individual annotator idiosyncrasy and algorithm performance. In general, we find that algorithms perform better on annotators who tend to agree with others, but that annotator response rate does not have a statistically significant positive correlation with algorithm performance.
\end{abstract}

\section{Introduction}

Perspectivist approaches to natural language processing (NLP) have recently increased in popularity \cite{fleisig_perspectivist_2024}, with a significant survey paper published by \citeauthor{frenda_perspectivist_2025} and multiple competitions measuring model performance on perspectivist tasks \cite{leonardelli_lewidi-2025_2025}. We use these English text-to-rating tasks as a wide testbed for observing the behavior of pluralist algorithms in order to understand what sorts of human behavior lead to poor predictions of beliefs under existing pluralist algorithms.

We believe our work would be of interest to these groups:
\begin{description}
\item[The first group] is AI alignment researchers who are interested in early detection of common behavior across a wide range of algorithms that indicates bias against (or inability to model) humans who exhibit certain types of behavior. Failure across different algorithm classes to accurately predict some people's opinions as well as others indicates the potential for pluralistic imputation models to fail to properly understand the values of certain people. We hope that this work can provide additional evidence that people with unpopular opinions may be more difficult for pluralistic systems to model.

\item[The second group] of people is privacy-sensitive individuals who are curious about the impact of withholding interaction-based data (e.g., ratings) from individualized prediction algorithms. While our experiments cover a relatively small niche of the space of individualized prediction algorithms, our results provide evidence that individuals who do not share large amounts of data will not suffer from algorithm performance decreases in comparison to individuals who do share large amounts of data.

\item[The third group] is creators of NLP datasets. In the process of data collection, a common concern is how to balance the time spent on building strong annotator guidelines that promote annotator agreement, versus focusing more intently on data collection despite disagreement. While many different factors may impact the ideal balance between those goals, we offer an insight from the perspective of model performance on individual annotators: the algorithms we tested tended to perform better for annotators who agree with one another, whereas the amount of data an annotator contributed to training data did not have a statistically significant correlation with algorithmic performance on that annotator.
\end{description}

Overall, we test the following two hypotheses across \NumDatasets{} datasets and \NumAlgos{} algorithms, after controlling for the number of annotations provided by an annotator in the test data:
\begin{enumerate}[label=(H\arabic*), ref=H\arabic*, leftmargin=*]
    \item \label{hyp:agreement} Lower individual annotator agreement is associated with worse model performance on that individual.
    
    \item \label{hyp:response-rate} Lower individual annotator response rate is associated with worse model performance on that individual.
\end{enumerate}

Our results show statistically significant evidence in support of \ref{hyp:agreement}, but do not support \ref{hyp:response-rate}.

\section{Related Work}

\paragraph{Perspectivist Approaches to NLP.}
Previous approaches to natural language processing datasets often involved using annotators to acquire an aggregated \QuoteUnquote{gold-label,} a single annotation that represented the correct answer for a natural language task \cite{aroyo_truth_2015, basile_its_2020}. Work on natural language processing has recently explored more perspectivist approaches, wherein individual annotators' perspectives are retained, and the goal of prediction tasks is to model individual annotations or distributional dynamics, rather than a single \QuoteUnquote{correct} label \cite{cabitza_toward_2023}.

Approaches related to perspectivism in natural language processing may use terminology such as \QuoteUnquote{subjectivity} \cite{homayounirad_will_2025}, \QuoteUnquote{uncertainty} \cite{zhang_disagreement_2025}, \QuoteUnquote{disagreement} \cite{davani_dealing_2022}, \QuoteUnquote{pluralistic} \cite{castricato_persona_2025, sorensen_position_2024}, or \QuoteUnquote{descriptive} approaches \cite{rottger_two_2022}.

The survey by \citeauthor{frenda_perspectivist_2025} provides a general overview of work in this area, highlighting the NLPerspectives workshop and Learning with Disagreements (LeWiDi) as core sources for research on this topic \cite{frenda_perspectivist_2025}. We build upon the work done by \citeauthor{lowmanstone_annotation_2023}, which was previously published in NLPerspectives, and add additional datasets from the SemEval 2023 LeWiDi competition \cite{lowmanstone_annotation_2023, leonardelli_semeval-2023_2023}.

Our research focuses primarily on individual modeling, as described by \citet{sorensen_value_2024}. While previous work has used demographic data to make individual and distributional predictions \citep{gordon_jury_2022, meister_benchmarking_2025}, due to concerns about using demographic data for human behavior prediction \citep{patty_algorithmic_2023, jernigan_gaydar_2009, barocas_big_2016-2}, we utilize algorithms that make predictions based on annotator behavior, rather than demographic information.

\paragraph{Pluralistic AI Alignment.}
Perspectivist  algorithms are a form of \QuoteUnquote{steerable pluralistic models} \citep{sorensen_position_2024}, which are trained to model diverse ranges of opinions. Among the goals of pluralistic AI alignment are to ensure that AI can follow the individual revealed preferences of people in diverse populations \citep{gabriel_artificial_2020}. We view NLP annotation prediction tasks as an early-stage test of AI systems' capability to model simple revealed preferences about text.

\paragraph{Fairness.}
Our work is also motivated by fairness in machine learning algorithms. Specifically, we are curious if annotators who respond less or differently from other annotators are modeled poorly by existing algorithms as a form of disparate impact \cite{pessach_review_2023}. Papers such as \citet{gordon_jury_2022} and \citet{eckman_aligning_2025} demonstrate that, for their respective tasks, underrepresented populations in training data lead to biased algorithmic performance, which may be rectified by amplifying the signal provided by underrepresented populations in the data.

Other work attempts to measure bias in existing NLP systems \citep{nangia_crows-pairs_2020, parrish_bbq_2022, nadeem_stereoset_2021, abid_persistent_2021, bolukbasi_man_2016}. We consider our work to be within this thread: instead of measuring bias against populations that are demographic minorities, we measure the general algorithmic bias against populations that are minorities in terms of opinions and response behavior.

\paragraph{Perspectivist Datasets.}
Multiple algorithms we use for annotation prediction make scalar predictions, so, in this work, we restrict our focus to datasets that use ordinal annotations (i.e., binary or rating-based annotations) rather than datasets with purely categorical annotations. However, many additional datasets have been released under the wider umbrella of perspectivist NLP annotation prediction.

The collection of perspectivist NLP datasets from \citet{plank_problem_2022} provides an excellent foundation for datasets with disaggregated annotations. To the best of our knowledge, the Awesome Human Label Variation (AHLV) repository provides the largest release of perspectivist datasets to date \citep{plank_problem_2022}. However, many of the datasets in AHLV are not NLP-specific, nor do they provide ordinal ratings. We use 4 of the datasets originally included in the collection, and 12 out of \NumDatasetFamilies{} of the dataset families we use in our analysis are currently listed in AHLV. We believe that our work will highlight additional NLP datasets that can be added to the repository.

\citet{deng_you_2023} provide a collection of datasets named TID-8 for disagreement-based tasks, \TIDDatasetCount{} of which are included in our collection. To the best of our knowledge, this is the previously largest released collection of solely perspectivist NLP annotation prediction datasets \cite{deng_you_2023}. In their paper, they demonstrate that modeling the underlying opinions and annotators that lead to a particular gold-label can improve models' ability to accurately predict individual labels. While the Pejorative \cite{dinu_computational_2021} dataset from TID-8 may be included in future work, other datasets, such as Friends QIA \cite{damgaard_ill_2021} and GoEmotions \cite{demszky_goemotions_2020}, are not suitable for our task, primarily due to a lack of clearly ordinal ratings.

Other tasks for NLP annotation prediction have been released via competitions. The LeWiDi 3 benchmark, released during the 2025 LeWiDi competition \cite{leonardelli_lewidi-2025_2025} includes additional perspectivist NLP datasets such as the Conversational Sarcasm Corpus \citep{jang_generalizable_2024}, MultiPICo \citep{casola_multipico_2024}, and Par \citep{leonardelli_lewidi-2025_2025}, which are not included in our datasets because their text does not consist of a single string; without a canonical method of combining the dataset example texts into a single string, we are concerned we may inadvertently bias algorithmic performance via our choice of format. Other datasets in the competition were not immediately suitable for our task because they utilized non-ordinal annotation categories.

\paragraph{Perspectivist Algorithms.}
Multiple types of algorithms can be used to accomplish perspectivist NLP tasks. Collaborative filtering approaches naturally apply but require additional mechanisms in order to ingest text \cite{su_survey_2009, ekstrand_collaborative_2011, shi_collaborative_2014}. General recommender system approaches may be applied as well, but are restricted to models that can accomplish rating prediction rather than item ranking \cite{ekstrand_collaborative_2011}. Matrix factorization approaches are also applicable to rating prediction tasks \cite{rendle_neural_2020}.

A recent wave of research has used LLMs to imitate human annotations, which can be viewed as a prediction task \cite{castricato_persona_2025, marreddy_usdc_2025}. However, existing studies show that LLM performance is highly task dependent, and previous models do not perform particularly well on NLP perspectivist tasks \cite{bavaresco_llms_2025, lowmanstone_annotation_2023, sarumi_corpus_2024}.

\paragraph{Metrics.}
Methods for measuring model performance on NLP perspectivist datasets have varied widely \cite{sarumi_nlp-resteam_2025}. Potential metrics include F1 scores, mean absolute distance, and ROC-AUC, which can also be measured on a per-annotator basis, rather than a per-dataset basis \cite{sarumi_nlp-resteam_2025, heinisch_architectural_2023}. For our experiment, we have chosen to use normalized mean squared error (NMSE) as our metric because NMSE is a natural error metric for ordinal ratings across multiple scales, compared to other metrics which are primarily used for categorical data and do not take advantage of the distance between ordinal ratings to measure the strength of poor predictions. While normalized root mean squared error (NRMSE) is a stronger choice for interpretability applications, we found that our statistical error analysis was more easily conducted using normalized mean squared error directly.

\begin{figure*}[t]
    \centering
    \includegraphics[width=0.9\textwidth]{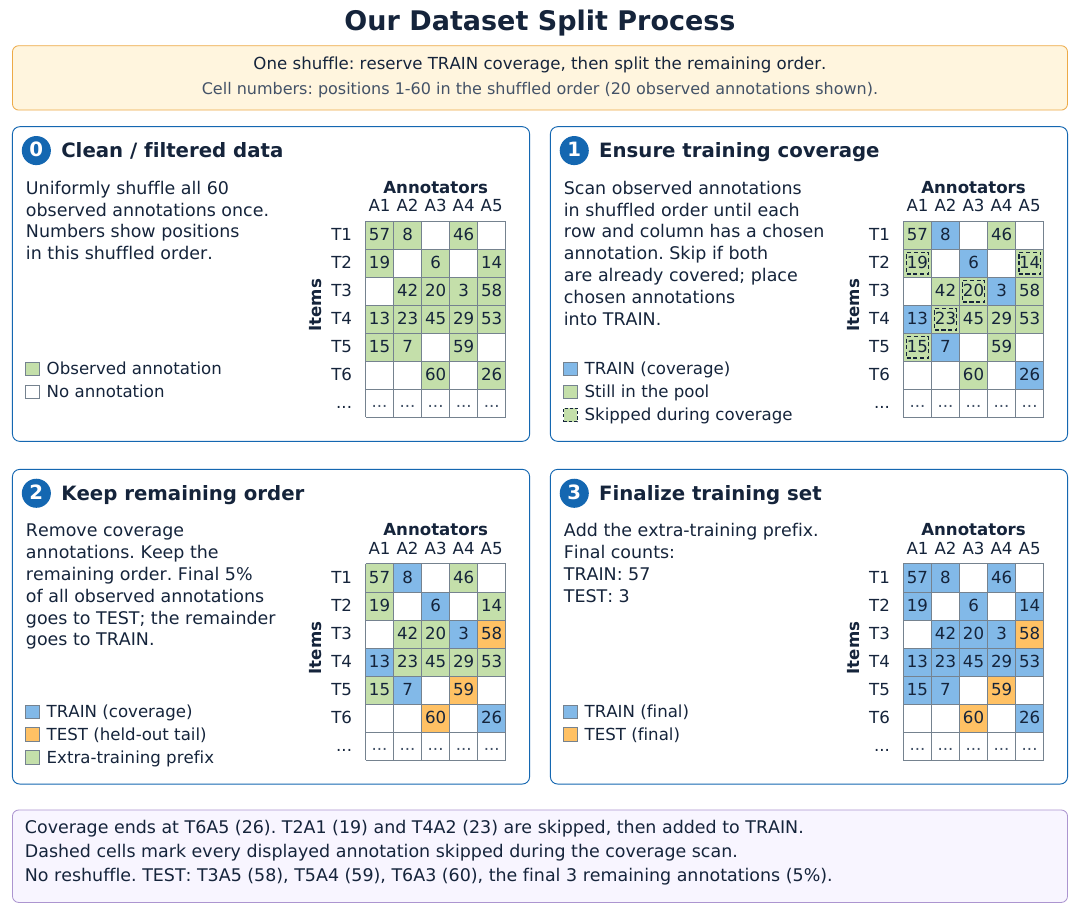}
    \caption{Depiction of the dataset split process. The figure displays 20 observed annotations from an illustrative pool of 60; cell numbers indicate positions in the shuffled order. The final test set contains the last 3 remaining annotations (5\%), and the training set contains 57 annotations.}
    \label{fig:dataset-split-process}
\end{figure*}

\section{Datasets}
\label{sec:datasets}

In our work, we use \NumDatasets{} English pluralist NLP datasets gathered from a variety of different sources, summarized in Table~\ref{tab:datasets} in the appendix. Information about licenses for each of the datasets can be found in Appendix~\ref{apx:dataset}. To the best of our knowledge, this is the largest collection of cleaned and formatted pluralist NLP datasets.\footnote{All data and code used in our experiments will be released (pursuant to licenses) upon paper acceptance.}

The datasets are drawn from AffectiveText
\citep{strapparava_semeval-2007_2007}; HuBi
\citep{kazienko_human-centered_2023}, including Cockamamie Gobbledegook
\citep{gultchin_humor_2019}, the Personal Bias emotion and valence datasets
\citep{milkowski_personal_2021}, and Wikipedia Detox
\citep{wulczyn_ex_2017-1}; NLPerspectives
\citep{lowmanstone_annotation_2023}, including SChem and SChem5Labels
\citep{forbes_social_2020}, SBIC
\citep{sap_social_2020}, GHC
\citep{kennedy_introducing_2022}, Sentiment
\citep{diaz_addressing_2018, diaz_age_2020}, and Politeness
\citep{danescu-niculescu-mizil_computational_2013}; LeWiDi
\citep{leonardelli_semeval-2023_2023}, including ConvAbuse
\citep{cercas_curry_convabuse_2021}, HS-Brexit
\citep{akhtar_whose_2021}, and MD-Agreement
\citep{leonardelli_agreeing_2021}; and additional datasets from AHLV \citep{plank_problem_2022}, including CommitmentBank \citep{marneffe_commitmentbank_2019}, Measuring Hate Speech \citep{sachdeva_measuring_2022}, Annotators
with Attitudes \citep{sap_annotators_2022}, and DynaSent
\citep{potts_dynasent_2021}.

Datasets were cleaned via a multi-step process. First, following the same dataset format as \citet{lowmanstone_annotation_2023}, they were converted from their original release files into two separate files: a matrix of nonnegative integer annotation values and a list of texts. Each text corresponds to a row in the matrix of annotations, and each annotator for the dataset corresponds to a column in the matrix. Ratings in datasets such as SBIC that originally allowed for 0.5 increments in ratings were doubled in order to create integer ratings. Missing annotations are represented by a unique sentinel value. Next, we merge duplicate texts in each dataset, keeping all annotations whenever possible, and choosing annotations at random if annotations conflict. Finally, to reduce the size of datasets in order to improve algorithmic speed, we uniformly sample 10,000 texts and 500 annotators independently and without replacement in order to create the final downsampled datasets. Examples and annotators without any annotations after downsampling are removed. For datasets with less than 10,000 texts or 500 annotators, all the data were kept.

After dataset cleaning, we split the datasets into train and test data using a 95\%-5\% split, as done by \citet{lowmanstone_annotation_2023}, which we detail here and depict in Figure~\ref{fig:dataset-split-process}, as it is particularly relevant to our experimental design. We first choose a random ordering of all annotations. Then, we iterate through this ordering, adding each annotation to the train split as long as the train split doesn't already contain annotations from its same row and column. We continue this process of adding annotations to the train split until there is at least one annotation for each row and each column in the train split. From the annotations not placed into the train split, we retain the original ordering, but iterate in reverse and assign 5\% of the total amount of annotations to create the test split. The remaining annotations (in the middle of the ordering) are added to the train split. This ensures that each text and annotator has at least one annotation during training, which is required for proper utilization of all of the chosen algorithms. This process introduces a dependency between the number of ratings provided by an annotator in the train split and the number of ratings provided by an annotator in the test split, which must be accounted for in our statistical analysis. Graphs of the distribution of annotator sparsity and agreement can be found in Figure~\ref{fig:annotator-data-dist} in the appendix.

\section{Algorithms}

We use \NumAlgoFamilies{} core families of imputation algorithms: 1. Neural Collaborative Filtering (NCF) \citep{he_neural_2017-1}, 
2. Multitask \citep{davani_dealing_2022}, 
3. Kernel Matrix Factorization (KMF) \citep{do_matrix_2022}, 
4. MASDA \citep{sullivan_university_2023},
5. TabFM \citep{kong_introducing_2026},  
and 6. baseline mean prediction algorithms, as listed in Table \ref{tab:algorithms} in the appendix.

 Across all models, if outputs fall outside the valid rating range for a given dataset, we clip the outputs into the proper range. For NCF and KMF, we follow the publicly released code from  \citet{lowmanstone_annotation_2023}, with only minimal adjustments for operating system differences. For Multitask, we introduce multiple architectural improvements, including adding checkpointing functionality and ensuring the number of classes output by the model is not greater than the number of possible classes in the dataset (see Appendix~\ref{apx:multitask-implementation}).

Unlike \citet{lowmanstone_annotation_2023}, we do not use ChatGPT or other LLMs to directly impute data; in our preliminary experiments, doing so was found to be cost-prohibitive and did not provide any performance gains over existing methods.

Further details about the Multitask improvements and implementation details about MASDA, TabFM, and the baseline mean algorithms can be found in Appendix~\ref{apx:algorithms}.

\section{Experiments}
\label{sec:experiments}

\subsection{Primary Hypothesis Testing}
To test \ref{hyp:agreement} and \ref{hyp:response-rate}, we train each of the \NumAlgos{} algorithms on each of the train splits, and then use each algorithm to predict each missing annotation in its training data. Each model is run once on each dataset. We can then compare the predictions made by the algorithm to the test data in order to compute performance.

Formally, let $m \in M$ indicate the algorithm (model) being used for prediction. Similarly, let $d \in D$ indicate the dataset and $a \in A_d$ indicate a particular annotator within the dataset.

We model algorithm performance by using weighted fixed-effects linear regression with cluster-robust standard errors, clustered by dataset-annotator pair and dataset-algorithm pair to account for residual dependence:
\begin{equation}
\label{eq:main-model}
Y_{mda} = \alpha_{md} + \beta_D Z^{D}_{da} + \beta_S Z^{S}_{da} + \gamma Z^{C}_{da} + \varepsilon_{mda}
\end{equation}
where $Y_{mda}$ is the normalized mean squared prediction error (NMSE) for algorithm $m$ on annotator $a$ in dataset $d$. The variable $\alpha_{md}$ represents a fixed effect parameter for dataset-algorithm pair $(d,m)$ to account for general model performance on datasets that is not related to specific annotators. $Z^{D}_{da}$, $Z^{S}_{da}$, $Z^{C}_{da}$ are normalized z-scores for the disagreement score, sparsity score, and log test count, respectively, of annotator $a$ in dataset $d$. $\beta_D$, $\beta_S$, and $\gamma$ are the coefficient parameters of the model, indicating the strength and directionality of the association between individual annotator disagreement, sparsity, and the amount of test annotations with respect to model performance. $\varepsilon_{mda}$ represents the error not accounted for by the model.

Let $r_{ida}$ be the rating (annotation) given by annotator $a$ on dataset $d$ for text $i$, and $\RMISS$ be the sentinel missing value. Furthermore, let $A'_d = A_d \setminus \{a\}$ represent all annotators other than annotator $a$. Finally, let $X^D_{ida}$ be the \textit{disagreement score} of annotator $a$ in dataset $d$ on text $i$ in the dataset. $X^D_{ida}$ is only defined for dataset row $i$ if target annotator $a$ and at least one other annotator have provided an annotation for text $i$. Then $$
X^D_{ida} = \frac{
    \sum_{a' \in A'_d} \mathbf{1}\!\left\{ r_{ida'} \neq \RMISS \land
    r_{ida'} \neq r_{ida} \right\}
}{
    \sum_{a' \in A'_d} \mathbf{1}\!\left\{ r_{ida'} \neq \RMISS \right\}
}$$
is the disagreement score for a given text $i$, dataset $d$, and target annotator $a$, computed as the proportion of annotators (not including annotator $a$) who disagreed with the target annotator on that text. The annotator's disagreement score $X^D_{da}$ on a dataset is then the average of their disagreement score over all of the rows {$i$} in the dataset for which $X^D_{ida}$ is defined, and $Z^D_{da}$ is the normalized z-score of $X^D_{da}$ across annotators in dataset $d$. We use the train split of dataset $d$ for this calculation. Annotators with an undefined disagreement score were not included in the regression.

We compute an annotator's \textit{sparsity} on a dataset by computing the proportion of texts that the annotator did not respond to in the dataset: $$X^S_{da} = \sum_{i \in I_d}\frac{\mathbf{1}\{r_{ida} = \RMISS \}}{|I_d|},$$ where $I_d$ is the set of row indexes for dataset $d$. $Z^{S}_{da}$ is then the normalized z-score of $X^{S}_{da}$ across annotators in dataset $d$. Again, the train split of dataset $d$ is used for this calculation.

\begin{table*}[ht]
\centering
\begin{tabular}{lrrrr}
\toprule
Hypothesis term & Coef. & 95\% CI & One-sided $p$ & Holm-adjusted $p$ \\
\midrule
Disagreement ($\beta_D$ on $Z^{D}_{da}$) & \textbf{0.025} & \textbf{[0.016, 0.033]} & $\mathbf{1.90 \times 10^{-8}}$ &
$\mathbf{3.80 \times 10^{-8}}$ \\
Sparsity ($\beta_S$ on $Z^{S}_{da}$) & 0.001 & [-0.001, 0.003] & 0.158 & 0.158 \\
\bottomrule
\end{tabular}
\caption{Results for the regression shown in Eq.~\ref{eq:main-model} with cluster-robust standard errors. Statistically significant results are in \textbf{bold}.}
\label{tab:main-results}
\end{table*}

The \textit{log test count} of an annotator is computed as $$X^C_{da} = \ln(1 + t_{da}),$$
where $t_{da}$ is the number of test annotations from annotator $a$ in the test split for dataset $d$. This term is included in the model due to the train/test split process shown in Figure~\ref{fig:dataset-split-process}: since annotations are sampled to cover each row and column in the train split, annotators with fewer examples are less likely to appear in the test set. This term controls for that effect, ensuring that $\beta_S$ better reflects effects from individual annotator sparsity in the train data, rather than a confounding relationship between the number of annotations provided by the annotator between the train and test splits. $Z^{C}_{da}$ is then the normalized z-score of $X^{C}_{da}$ across annotators in the test split of dataset $d$.

We fit the model via weighted least squares, with the regression weight being the number of held-out test annotations $t_{da}$ divided by the number of algorithms in that algorithm's family. This gives greater influence to annotator-level NMSE estimates based on more test annotations, which are expected to be more stable than estimates computed from fewer held-out annotations.

To compute confidence intervals and p-values for our estimates of $\beta_D$ and $\beta_S$, we use cluster-robust standard errors clustered by dataset–annotator pair and dataset-algorithm pair (in order to adjust for correlated residuals across algorithms and annotators, respectively) and apply a standard small-sample correction.

Formally, the two main hypotheses we are testing are  $\beta_D > 0$ (\ref{hyp:agreement}) and $\beta_S > 0$ (\ref{hyp:response-rate}). Since we test two hypotheses, we adjust the resulting p-values, computed via Wald $z$-tests, using the Holm–Bonferroni correction. We report 95\% two-sided confidence intervals for $\beta_D$ and $\beta_S$ and the one-sided and Holm-adjusted p-values for each of the hypothesis tests. A positive coefficient indicates a positive relationship between the corresponding z-score and NMSE (a measure of error); as the variable increases, algorithm performance decreases.

\subsection{Post Hoc Analysis}
In addition to our main hypothesis testing, we analyze additional data to ensure that no particular dataset or algorithm family is heavily skewing our results, as such an outcome would limit the generalizability of our claims. We emphasize here that no models were removed from our hypothesis testing or results as a consequence of these post hoc analyses.

To determine the impact of dataset families on our primary result, we conducted bootstrap analyses with 10,000 samples each. There are \NumDatasetFamilies{} different dataset families, as specified by the row groupings in Table~\ref{tab:datasets} in the appendix. Within each bootstrap sample, we sample \NumDatasetFamilies{} dataset families with replacement from the set of possible dataset families. Then, we refit our statistical model from Eq.~\ref{eq:main-model} using the data from those sampled families (where families chosen multiple times during the sampling step had all the data from the datasets of that family included as many times as that family was sampled). This enables us to estimate how our results may have shifted if certain families of datasets were not included or were overly represented. The resulting plots are in Figure~\ref{fig:dataset-family-block-bootstrap}.

Since the number of algorithm families is small, with \NumAlgoFamilies{} groups of related algorithms, the fidelity of bootstrapping is expected to drop significantly if applied to algorithm families alone \cite{cameron_bootstrap-based_2008}. Leave-one-out methods are more appropriate for indicating algorithm family impact on regression coefficients. Thus, we re-fit the model from Eq.~\ref{eq:main-model} \NumAlgoFamilies{} times, leaving each algorithm family out exactly once. The results are as a forest plot in Figure~\ref{fig:leave-one-algo-family-out}.

In Appendix~\ref{apx:post-hoc-analysis}, we provide heatmaps of algorithm family and individual algorithm performance on each dataset.

\section{Results and Analysis}

After fitting the model on our data and computing the error estimation, we obtain the following results listed in Table~\ref{tab:main-results}.

The results of our experiment indicate that it is highly likely that individual annotator agreement is positively associated with algorithm performance; that is, individual annotator disagreement is positively correlated with increased NMSE. More specifically, our evidence indicates that, when compared to other annotators within the same dataset, each standard deviation of increased disagreement score from an individual annotator corresponds to an increase of over 0.02 (on average) in the algorithms' NMSE on that annotator.

Since NMSE is normalized by the possible rating range of a dataset, this indicates a shift in MSE above 2\% of the squared possible range of annotations. For example, on datasets with a range of 100, this indicates an average shift of over 200 in the MSE for each standard deviation that an annotator's disagreement score is higher than the mean disagreement score for that dataset.

Our primary results (see the Sparsity effect row of Table~\ref{tab:main-results}) show no significant evidence of a positive association between response rate and algorithm performance. This is a fairly unintuitive result, as our initial hypothesis proposed that less training data on a particular annotator would lead to worse performance. However, it appears that for our tasks and algorithms, agreement is far more predictive of algorithm performance than response rate, and response rate may even have performance associations that are counter to our initial hypothesis.

\begin{figure*}[t]
    \centering
    
\includegraphics[width=0.49\linewidth]{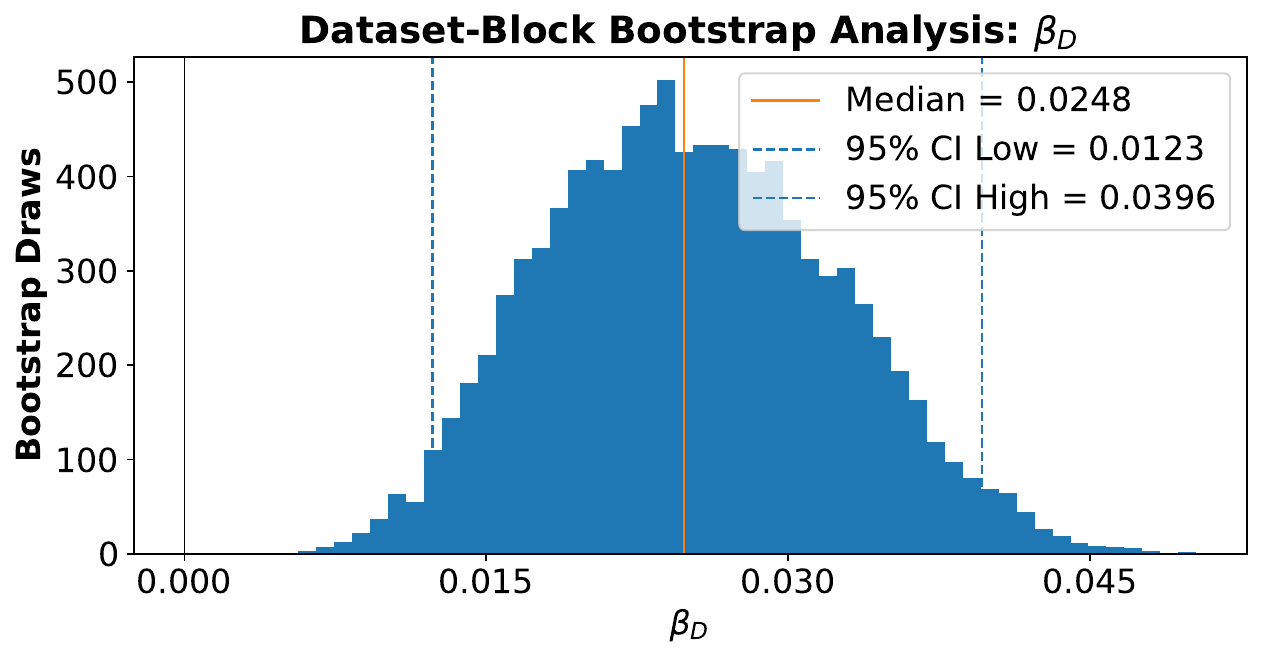}
    \hfill
    \includegraphics[width=0.49\linewidth]{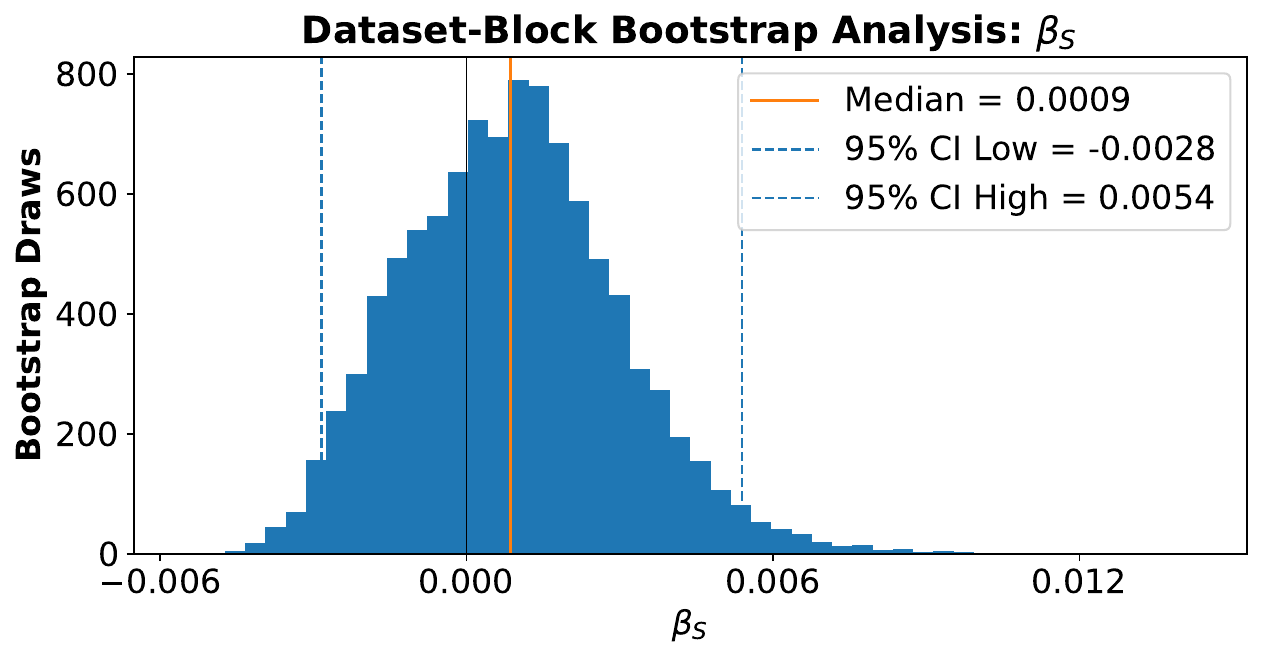}
    
    \vspace{0.85em}

    \begin{tabular}{lrrr}
      \hline
      \textbf{Coeff.} & \textbf{Median} & \textbf{95\% CI} & \textbf{Pos. Coeff. Rate} \\
      \hline
      $\beta_D$ & 0.0248 & [0.0123, 0.0396] & 1.0000 \\
      $\beta_S$ & 0.0009 & [-0.0028, 0.0054] & 0.6554 \\
      \hline
    \end{tabular}

    \caption{Dataset-family block bootstrap results from 10,000 samples each. Top: histograms displaying the estimated regression slope across samples. Left: disagreement effect. Right: sparsity effect. Bottom: 95\% bootstrap confidence intervals and positive coefficient rates. Positive coefficient rate denotes the proportion of bootstrap samples in which the coefficient estimate was positive. The orange vertical line highlights the median value, and the black vertical line highlights a value of 0. Our hypotheses predicted that both distributions would lie significantly to the right of the black line.}
    \label{fig:dataset-family-block-bootstrap}
\end{figure*}

\begin{figure*}[ht]
  \centering
  \includegraphics[width=0.49\linewidth]{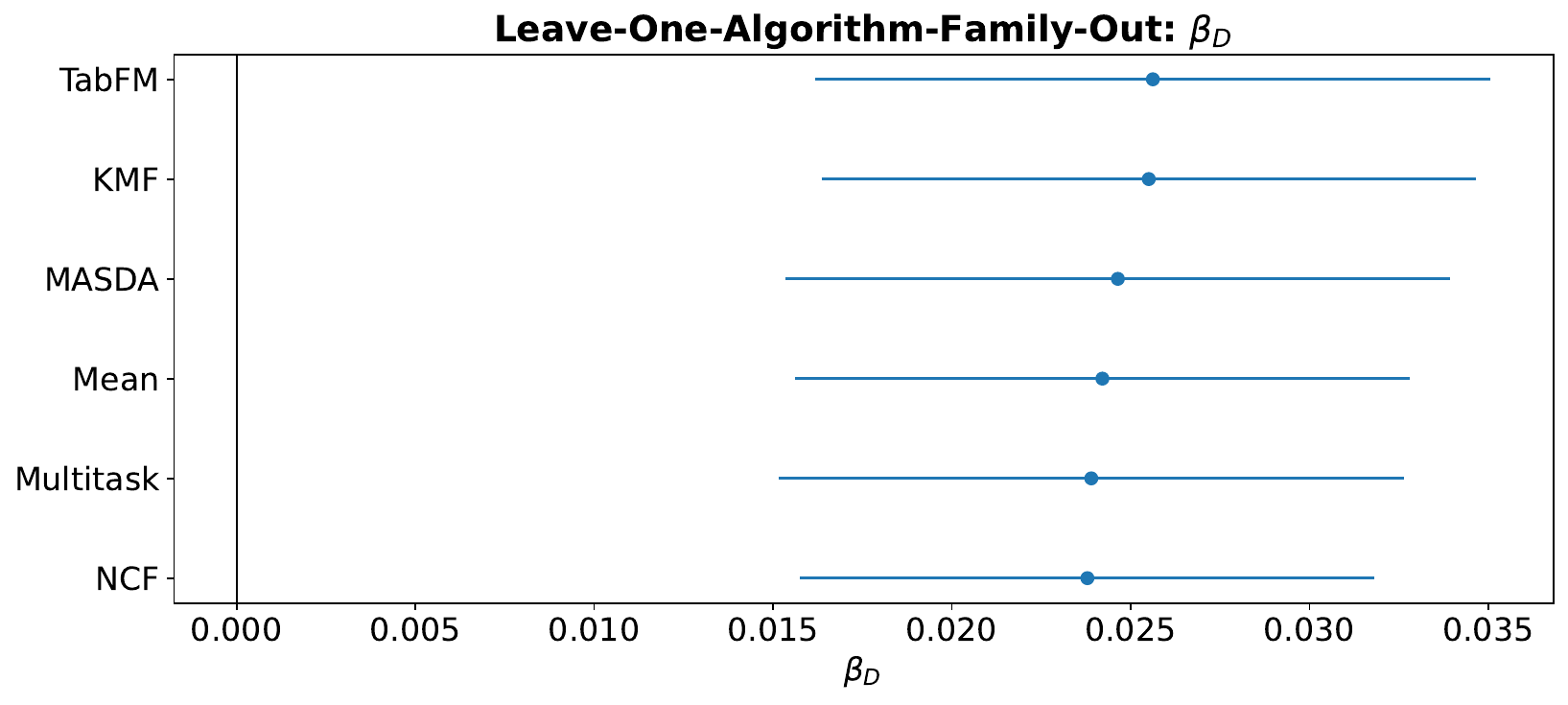}
  \hfill
  \includegraphics[width=0.49\linewidth]{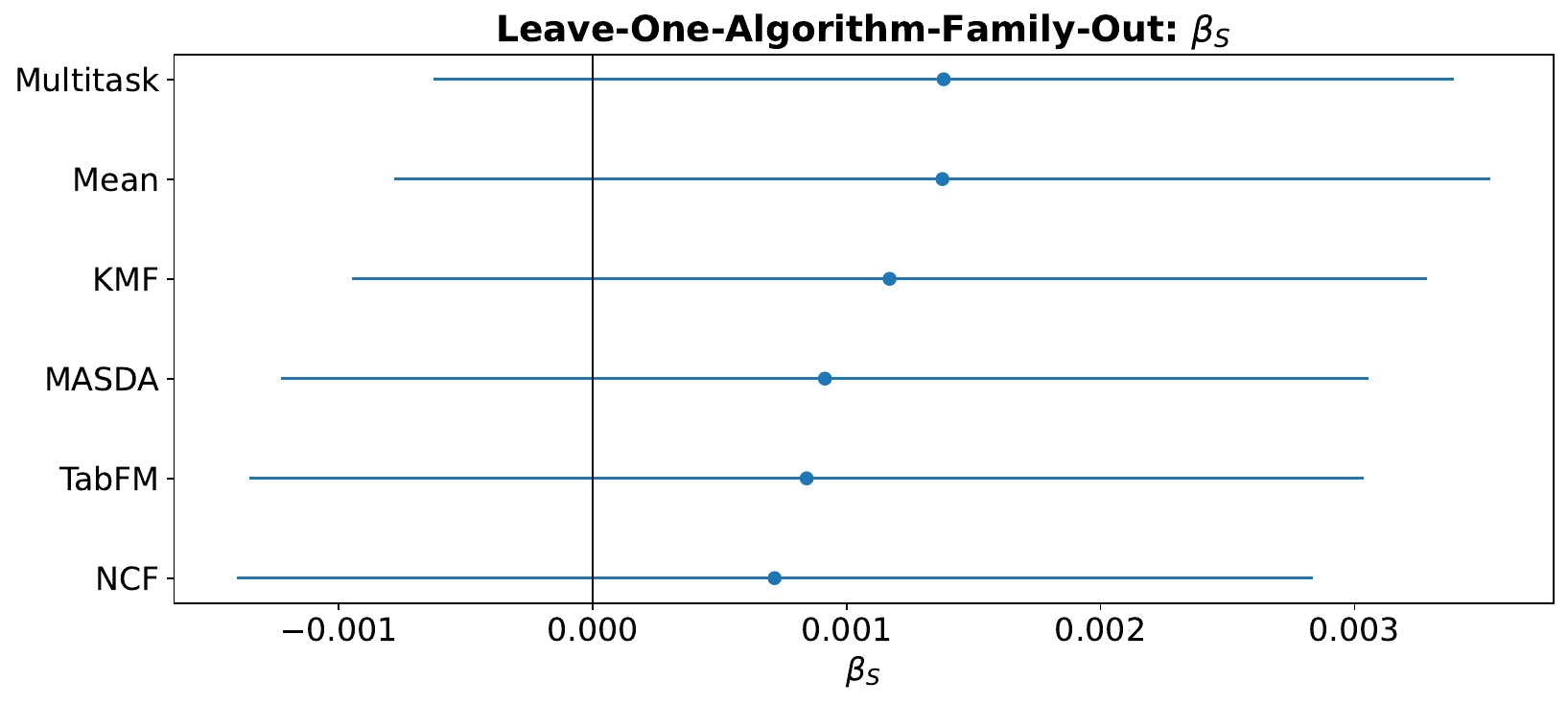}
  \caption{Forest plot from fitting the regression model from Equation~\ref{eq:main-model} with one family of algorithms left out of the analysis. The figure displays the disagreement effect (left) and sparsity effect (right). Error bars indicate the 95\% CI.}
  \label{fig:leave-one-algo-family-out}
\end{figure*}

The post-hoc analysis in Figure~\ref{fig:dataset-family-block-bootstrap} demonstrates that while specific dataset families have some impact on the disagreement coefficient during regression, the tested algorithms tend to display a statistically significant correlation between annotator disagreement and performance across all datasets. 

With regard to sparsity, we observe that, bootstrapping with dataset-family blocks roughly mirrors our main results from Table~\ref{tab:main-results}, showing a statistically insignificant negative relationship between annotator sparsity and algorithm performance. It is worth noting that the 95\% confidence interval contains 0, indicating that the direction of the response-rate association with algorithm performance may be fairly dataset-dependent.

With regard to algorithm families, Figure~\ref{fig:leave-one-algo-family-out} demonstrates that no individual algorithm family significantly impacts the results for either of $\beta_D$ and $\beta_S$.

\section{Conclusions}

Overall, we find strong evidence supporting that individual annotator agreement has a positive correlation with algorithm performance for that annotator, but find that the number of annotations provided by an annotator during training does not have a statistically significant positive association with algorithm performance. For AI alignment researchers, this provides evidence that existing algorithms for pluralist prediction of human behavior regarding opinions of text tend to be biased towards humans with popular opinions; more alignment work may be needed to ensure that future algorithms are aligned for people with unpopular opinions as well. For dataset creators who want high model performance, this provides evidence towards focusing on annotator guidelines in order to increase agreement. For privacy-sensitive individuals, our results suggest that withholding additional data is unlikely to significantly harm performance for these personalized algorithms, though the correlational nature of our experiments cannot confirm this.

\section*{Limitations}
\label{sec:limitations}
While our work does provide, to the best of our knowledge, the largest collection to date of perspectivist datasets with ordinal annotations, it is not an exhaustive set. We also restrict our domain to English-only datasets since the Multitask and TabFM algorithms rely on English-based text encoders. Future work may include the addition of non-English datasets and text encoders.

Similarly, while we test multiple model architectures, prediction methods, and encoder implementations, there are many approaches to making individualized predictions on perspectivist NLP datasets that are not clearly covered by our selection of algorithms.

In the interpretation of our results, we rely on the original creators of the datasets to have implemented proper safeguards to ensure that annotations faithfully represent annotators' true judgments of text. If the source datasets contain invalid annotations, that may impact the interpretation of our experimental data.

In addition, our study is correlational, not interventional; if there are unmeasured factors that jointly impact annotator sparsity and rating behavior, then our conclusion regarding data-sensitive individuals may not be correct, as we do not directly measure how models perform when training data is directly removed. Similarly, we do not directly test increasing annotator agreement across datasets, so any unobserved sources of association may weaken our conclusion regarding suggestions for dataset creators.

Initially, we hypothesized that higher annotator response rate would be positively associated with algorithm performance on those annotators. Even after controlling for the relationship between the number of train and test examples in our regression model, the evidence collected did not support our initial hypothesis. In order to retain scientific validity, we report the results of our investigation as-is, but further research may provide additional insights.

Overall, we tested \NumAlgos{} different algorithms split across \NumAlgoFamilies{} distinct algorithm families that are capable of making annotation predictions on a wide variety of perspectivist NLP datasets. However, our results may not generalize beyond these particular algorithms and datasets, and dataset collection techniques may change the interpretation of our results. We encourage other researchers and practitioners in this space to utilize our code and techniques to measure algorithmic sensitivity to annotator idiosyncrasy across their own unique datasets and algorithms and inform us of any conflicting results.

\section*{Acknowledgments}
A significant portion of experiments were conducted on \ComputeAcknowledgment.

AI assistants were used significantly to help implement experiments, adapt models to different hardware, and suggest and critique aspects of the experimental design and interpretation. All coding tasks were explicitly overseen by humans. Output code was either reviewed by humans or validated through human-designed and human-verified unit and integration tests.

AI assistants were used lightly during writing in order to provide suggestions for small individual phrases, correct grammar, create tables, and implement Appendix~\ref{apx:MASDA-variation-implementation}. Figure~\ref{fig:dataset-split-process} was also generated via AI systems. The authors bear full responsibility for any errors that appear in the final work.

\bibliography{arXiv-London}

\clearpage
\appendix

\section{Supplementary Algorithm Information}
\label{apx:algorithms}

The algorithms used, their corresponding family, and parameter counts are listed in Table \ref{tab:algorithms}. Model parameter counts vary per dataset with the amount of annotators and annotation value range; we list the maximum number of parameters of each model across all training sessions in the corresponding column.

Total estimated GPU use for these experiments is approximately 76.81 hours, with roughly 48.87 hours of computation taking place on A100 GPUs and 27.94 hours taking place on H100 GPUs.

\begin{table}[ht!]
    \centering
    \small
    \begin{tabular}{@{}llr@{}}
        \toprule
        \textbf{Algorithm} & \textbf{Family} & \textbf{\#Param} \\
        \midrule
        Example Mean & Mean & 0 \\
        Global Mean & Mean & 0 \\
        User Mean & Mean & 0 \\
        \midrule
        KMF & KMF & 672,815 \\
        \midrule
        MASDA-Categorical & MASDA & 23,868,264 \\
        MASDA-Ordinal & MASDA & 23,868,222 \\
        \midrule
        Multitask-ALBERT & Multitask & 15,893,859 \\
        Multitask-BERT & Multitask & 113,692,515 \\
        Multitask-DistilBERT & Multitask & 70,573,155 \\
        Multitask-GPT-2 & Multitask & 128,650,083 \\
        Multitask-RoBERTa & Multitask & 128,855,907 \\
        \midrule
        TabFM-Column & TabFM & 1,647,782,989 \\
        TabFM-Column+BERT & TabFM & 1,757,265,229 \\
        TabFM-Cell & TabFM & 1,647,782,989 \\
        \midrule
        NCF & NCF & 7,313,921 \\
        \bottomrule
    \end{tabular}
    \caption{Algorithms used in the experiments, grouped by source family, with maximum parameter counts across the experiment datasets.}
    \label{tab:algorithms}
\end{table}

\subsection{Multitask Implementation}
\label{apx:multitask-implementation}
We improve upon the Multitask model provided by \citet{lowmanstone_annotation_2023} in a few key areas. First, we implement model checkpointing in order to reduce overfitting during model training. The publicly released code base from \citet{lowmanstone_annotation_2023} utilized a 10\% internal validation set for monitoring and logging during training Multitask models. This validation set was created using the same process illustrated in the main paper in Figure~\ref{fig:dataset-split-process}. We implemented a checkpoint mechanism to fully utilize this validation data; the final prediction model was chosen to be the model with the lowest loss on the validation data, as measured after each of \MultitaskEpochs{} training epochs. This behavior differs from \citet{lowmanstone_annotation_2023}, which used the model trained for a fixed 10 epochs for prediction regardless of validation performance.

Next, in addition to the single BERT-base-uncased encoder, we test multiple different text encoding models, which are detailed in Table \ref{tab:algorithms}. Due to architectural differences in encoding, models from the BERT family (BERT \citep{devlin_bert_2019-1}, ALBERT \citep{lan_albert_2019-2}, DistilBERT \citep{sanh_distilbert_2019}, and RoBERTa \citep{liu_roberta_2019}) pass the representation of the first token in the sequence (i.e. the classification token) to the rest of the model, whereas the GPT-2 encoder \citep{radford_language_2019} passes the representation of the last non-padding token. Since we utilize multiple encoders with various properties, the context window was adjusted to the maximum size for each model, rather than being fixed at 512. To reduce memory usage, we also implemented an adaptive padding mechanism which pads each example up to the length of the longest example in its batch, rather than the full context length.

Finally, we adjusted the output range of the Multitask algorithm to match the training data. The original implementation from \citet{lowmanstone_annotation_2023} assumed that all dataset ranges had a minimum annotation value of 0 and used an architecture that could output annotation predictions of 0 even on datasets whose lowest annotation was above 0. We modified the architecture by no longer including extraneous output units that corresponded to those unseen annotations in the training data.

\subsection{MASDA Implementation}
\label{apx:MASDA-variation-implementation}

We implemented the MASDA model from \citet{sullivan_university_2023} as a prediction algorithm for our datasets. The original MASDA code was implemented for usage on binary datasets only. Since our datasets have ordinal ratings that are not binary, we implemented two different variations on the MASDA architecture to adapt it to our particular task. The first variation, MASDA-Categorical, replaces the original scalar logit decoder with a class logit decoder and exchanges the original binary cross-entropy loss for standard cross-entropy. The MASDA-Ordinal model predicts normalized ratings using the sigmoid of the scalar decoder output and computes MSE loss against labels normalized to the unit interval; predictions are then mapped back to the dataset-specific annotation scale. We describe both architectures below in detail.

\paragraph{Shared MASDA architecture.}
The MASDA architecture from \citet{sullivan_university_2023} consists of a shared text encoder, annotator-specific latent networks, a shared decoder, and an optional gold-label prediction pathway. We implemented two adaptations of this architecture for our ordinal annotation prediction setting: MASDA-Ordinal and MASDA-Categorical. We describe both using a shared template.

Let text \(i\) have input \(x_i\), and let \(r_{ij}\) be the rating provided by annotator \(j\). Let missing ratings be marked by \(\RMISS\), and define the observed-cell indicator
\[
o_{ij} = \mathbf{1}[r_{ij} \neq \RMISS].
\]
The text encoder first produces a text representation:
\[
H_i = \mathrm{LM}(x_i), \qquad h_i = \mathrm{MeanPool}(H_i).
\]
Each annotator-specific network then maps this text representation into an annotator-specific latent representation:
\[
u_{ij} = F_j(h_i),
\]
where \(F_j\) is the network associated with annotator \(j\). The shared decoder maps this latent representation into either a scalar logit or a vector of class logits, depending on the MASDA variant:
\[
z_{ij} = \mathrm{Dec}_\theta(u_{ij}).
\]
For variants that use the gold-label pathway, the model also computes attention weights over annotators,
\[
\omega_i = \mathrm{softmax}(W_g h_i),
\]
forms a weighted latent representation,
\[
g_i = \sum_j \omega_{ij}u_{ij},
\]
and decodes it:
\[
z_i^{(g)} = \mathrm{Dec}_\theta(g_i).
\]

During training, we detach \(h_i\) and each \(u_{ij}\) before passing them to the gold-label pathway. Consequently, the gold-label loss updates the attention module and shared decoder, but its gradients do not flow back into the text encoder or annotator-specific networks.

\paragraph{MASDA-Ordinal.}
MASDA-Ordinal preserves the scalar-logit decoder used by the original MASDA implementation, but interprets the decoder output as a position on the observed annotation scale rather than as a binary-classification logit. Let
\begin{equation*}
\begin{aligned}
    r_{\min} &= \min\{r_{ij}: o_{ij}=1\}, \\
    r_{\max} &= \max\{r_{ij}: o_{ij}=1\}, \text{ and} \\
    s &= r_{\max} - r_{\min}.
\end{aligned}
\end{equation*}

Observed ratings are normalized to the unit interval:
\[
\widetilde{r}_{ij} = \frac{r_{ij}-r_{\min}}{s}.
\]
The model predicts a normalized rating using the sigmoid of the scalar decoder output:
\[
\widehat{\widetilde{r}}_{ij} = \sigma(z_{ij}),
\]
and maps it back to the dataset-specific rating scale:
\[
\widehat{r}_{ij} = r_{\min} + s\widehat{\widetilde{r}}_{ij}.
\]
The MASDA-Ordinal loss is masked mean squared error on the normalized labels:
\[
L_{\mathrm{ord}}
=
\frac{1}{\sum_{i,j} o_{ij}}
\sum_i\sum_j
o_{ij}
\left(\sigma(z_{ij})-\widetilde{r}_{ij}\right)^2.
\]
The gold-label pathway is not used for MASDA-Ordinal. During imputation, observed cells retain their original ratings and missing cells are filled with the model prediction:
\[
r^{\mathrm{imp}}_{ij}
=
\begin{cases}
r_{ij}, & o_{ij}=1,\\
\widehat{r}_{ij}, & o_{ij}=0.
\end{cases}
\]

\paragraph{MASDA-Categorical.}
MASDA-Categorical instead treats each rating value as a distinct class. For observed ratings (\(o_{ij}=1\)), define the zero-based class index \(c_{ij}=r_{ij}-r_{\min}\in\{0,\dots,K-1\}\). The decoder therefore outputs a vector of class logits:
\[
z_{ij} \in \mathbb{R}^{K}, \qquad z_i^{(g)} \in \mathbb{R}^{K},
\]
where \(K\) is the number of classes. The predicted rating is
\[
\widehat{r}_{ij} = r_{\min} + \arg\max_{c \in \{0,\dots,K-1\}} z_{ij,c}.
\]
The gold-label target for the gold pathway is the row plurality class:
\[
t_i^{(g)}
=
\arg\max_c
\sum_{j:o_{ij}=1}\mathbf{1}[c_{ij}=c].
\]
The annotator-specific loss is masked cross-entropy:
\[
L_{\mathrm{ann}}
=
\frac{1}{\sum_{i,j} o_{ij}}
\sum_{i,j:o_{ij}=1}\mathrm{CE}(z_{ij}, c_{ij}).
\]
Let \(I_{\mathrm{obs}}=\{i:\exists j,\;o_{ij}=1\}\) be the set of rows with at least one observed annotation, and let \(N_{\mathrm{obs}}=|I_{\mathrm{obs}}|\). The gold-label loss is
\[
L_{\mathrm{gold}}
=
\frac{1}{N_{\mathrm{obs}}}
\sum_{i\in I_{\mathrm{obs}}}
\mathrm{CE}(z_i^{(g)}, t_i^{(g)}).
\]
The total MASDA-Categorical loss is
\[
L_{\mathrm{cat}}
=
L_{\mathrm{ann}}
+
\lambda_{\mathrm{gold}}L_{\mathrm{gold}},
\]
where \(\lambda_{\mathrm{gold}}\) is the gold-loss weight; in our experiments, we use the default value \(\lambda_{\mathrm{gold}}=1.0\). As with MASDA-Ordinal, imputation preserves observed ratings and fills missing cells with the model prediction:
\[
r^{\mathrm{imp}}_{ij}
=
\begin{cases}
r_{ij}, & o_{ij}=1,\\
\widehat{r}_{ij}, & o_{ij}=0.
\end{cases}
\]

\subsection{Performance Comparison}
\label{apx:MASDA-variation-perf-comp}

\begin{figure}[ht!]
    \centering
    \includegraphics[width=\columnwidth]{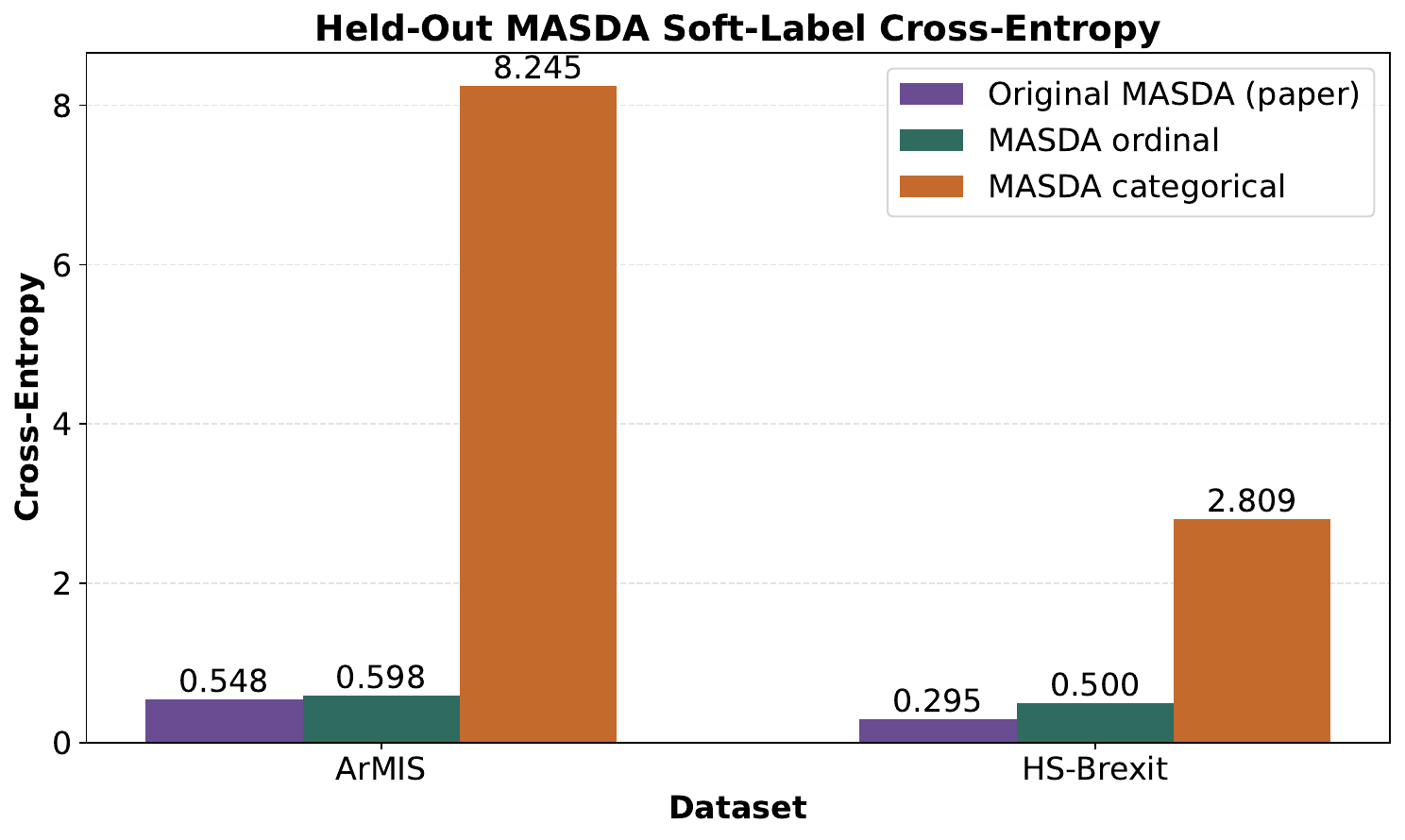}
    \caption{Soft-label cross-entropy scores of the original MASDA model and our ordinal and categorical MASDA variants on the ArMIS \citep{almanea_armis_2022} and HS-Brexit \citep{akhtar_whose_2021} datasets. Lower scores indicate better performance. Scores for our variants were computed on held-out annotations, while the original MASDA scores are those reported in the original paper and were not recomputed using our train/test splits.}
    \label{fig:MASDA-comparison}
\end{figure}

The performance of the two MASDA model implementations is compared against the original results reported by \citeauthor{sullivan_university_2023} in Figure~\ref{fig:MASDA-comparison}. Note that this is not a direct comparison, as the cross-entropy loss on models implemented by this work are measured on the test data derived from this work, which does not directly match the data the original model was tested on.

Both variations suffered noticeable losses of performance in comparison to the original MASDA implementation, which may be attributable to our unique evaluation environment. However, these additional models provide useful evidence that our analysis holds across multiple different algorithmic approaches to individualized prediction.

We include MASDA-Categorical because, despite its poor performance on cross-entropy, Figure~\ref{fig:algo-dataset-heatmap} demonstrates that it can outperform MASDA-Ordinal on a select few datasets. However, the MASDA-Categorical model can suffer from prediction collapse and be penalized highly as seen in Figure~\ref{fig:algo-dataset-heatmap}, especially on datasets such as AffectiveText ANGER, which has a highly skewed annotation distribution (as observed from Figure~\ref{fig:annotation-dist}) despite having a wide range of possible annotation values (see Table~\ref{tab:datasets}).

\subsection{TabFM Model Implementation}
The TabFM model from \citet{kong_introducing_2026} is a recently-released imputation method for tabular data. As a general description, the TabFM algorithm has two phases: fit, and predict.

\paragraph{Fit Phase.} During the fit phase, the model is not fine-tuned. Instead it processes the given data in order to feed it to a fixed neural architecture during the predict stage. The fit stage expects two inputs: 1. A Training Matrix (\texttt{X\_context}), which is a matrix where each row corresponds to an example, and each column contains features. 2. A Target Array (\texttt{y\_context}), which is a vector with one element for each row in the train context; each element is the target prediction for its corresponding row in the Training Matrix.

\paragraph{Predict Phase.}
The predict phase takes as input a Prediction Matrix (\texttt{X\_predict}) with the same number of columns as the Training Matrix; each row corresponds to asking the model to make a prediction given the features in that row. The model will then provide a Prediction Array output with one element for each row in the Prediction Matrix.

\paragraph{Additional Information.}
In both phases, missing feature entries in the Training Matrix and Prediction Matrix are, by default, automatically replaced with the mean of existing values from the Training Matrix for that feature.

In some scenarios (8.67\% of predictions for TabFM-Column and 1.81\% of predictions for TabFM-Column+BERT), the input features do not vary across rows in the Training Matrix. This can occur either when an annotator only has one example in the train split or when all other annotators who annotated the same examples as the target annotator gave the exact same annotations for each shared example. In these scenarios, we conduct a \QuoteUnquote{mean fallback}: instead of running TabFM (which would error on such an input), we instead predict the mean of the Target Array.

We use regression mode with all the default hyperparameters from the released TabFM code, with no hyperparameter search. Conducting a hyperparameter search is likely to improve performance. However, in order to conduct such a search, we either need to reserve data for validation which will not be used for training or overfit the TabFM model to our suite of datasets, reducing the value of our experimental analysis. Given these tradeoffs and that the models performed relatively well (see Figure \ref{fig:algo-performance}), we have currently opted to use the default hyperparameter settings.

\paragraph{TabFM-Column.}
For TabFM-Column, we create one Training Matrix for each unique annotator $a$ in the test split. This Training Matrix is constructed by first removing all rows from the train split matrix (assuming no text column) in which annotator $a$ does not provide an annotation. We then remove the column corresponding to $a$ from the resulting matrix and use that removed column as the Target Array. The remaining matrix is passed as the Training Matrix. The Prediction Matrix is created similarly, by removing all rows for which no prediction for annotator $a$ is requested. Then, the column corresponding to annotator $a$ is removed, and the resulting matrix is passed as the Prediction Matrix. TabFM will then make a prediction for annotator $a$ for each row in the Prediction Matrix. We repeat this process for each unique annotator in the test split.

\paragraph{TabFM-Column+BERT.}
The TabFM-Column+BERT model is identical to the TabFM-Column implementation, except for that we add extra features to the train split data. Before removing the text column, we encode each of the texts with BERT-base-uncased, resulting in a 768-dimensional CLS vector for each row. We then add 768 columns to each row, each column corresponding to a component of the text embedding from BERT. The rest of the process remains identical. This provides textual features to the TabFM model, enabling it to make predictions based on the textual data. Since these features will vary over multiple examples, the mean fallback was only used for annotators whose training data consisted of a single example. It is worth noting that the default TabFM implementation which we use samples only 500 features at a time. This low upper limit on the number of features may limit the viability of this approach, and may introduce a tradeoff between using text and other annotators' ratings in order to predict a target annotator's ratings.

\paragraph{TabFM-Cell.}
For this implementation, instead of each annotator being the main unit, each cell (rating) within the matrix is the main unit for data transformation. To create the Training Matrix we build one row for each rating $r_{ia}$ in the train split, where $i$ indexes the example and $a$ indexes the annotator. The row in the Training Matrix corresponding to rating $r_{ia}$  consists of the original row $i$ from the train split with a few modifications. We remove the rating $r_{ia}$ from the specified row $i$ and place it into the Target Array, replacing it with the missing sentinel value. Then, we remove the text column from the row, and prepend the numeric ID of annotator $a$ to the row as an additional feature column. Thus, the Training Matrix will have rows equal to the number of ratings in the training data, and columns equal to the number of annotators plus one (for the ID of annotator $a$). The Target Array will have one element for each rating in the training data. The Prediction Matrix is constructed analogously to the Training Matrix, except the step of removing the rating and placing it into a Target Array is no longer needed, since the rating is unknown (to the algorithm) by definition.

\subsection{Mean Prediction Algorithms Implementation}
To further expand algorithmic diversity, we also include baseline mean prediction algorithms. User Mean predicts the average annotation of each column in the train dataset, Example Mean predicts the average annotation of each row in the train dataset, and Global Mean predicts the average value across all annotations in the train dataset. While most algorithm families outperform this family of baseline mean algorithms (see Figure~\ref{fig:algo-performance}), we find that mean algorithms (especially Example Mean) can outperform all other algorithms on particular datasets, especially when measured using normalized root mean squared error (NRMSE), as shown in Appendix \ref{apx:algo-performance}.

\section{Supplementary Dataset Information}
\label{apx:dataset}

\begin{table*}[ht!]
    \centering
    \small
    \setlength{\tabcolsep}{4pt}
    \renewcommand{\arraystretch}{1.05}
    \begin{tabular}{L{2.7cm} L{2.8cm} r r C{1.5cm} L{5.5cm}}
    \toprule
    \textbf{Dataset} & \textbf{Source} & \textbf{N} & \textbf{M}
    & \makecell{\textbf{Ratings/}\\\textbf{Example}}
    & \textbf{Label Meaning} \\
    \midrule

    \makecell[l]{(Other) $\times$ 6 \\ VALENCE}
    & AffectiveText
    & \makecell[c]{987 \\ 987}
    & \makecell[c]{6 \\ 6}
    & \makecell[c]{6.0 \\ 6.0}
    & \makecell[l]{
        Scale from absent (0) to maximum (100) \\
        Highly negative (0), Highly positive (200)
    } \\
    \midrule

    Cockamamie Gobbledegook
    & HuBi
    & 10000
    & 351
    & 3.6
    & Not funny (0), Funny (1) \\
    \midrule

    \makecell[l]{
        (Other) $\times$ 8 \\
        EMOT. AROUSAL \\
        VALENCE
    }
    & \makecell[l]{HuBi/\\Personal Bias}
    & \makecell[c]{100 \\ 100 \\ 100}
    & \makecell[c]{500 \\ 500 \\ 500}
    & \makecell[c]{8.5 \\ 8.5 \\ 8.5}
    & \makecell[l]{
        Scale from low (3) to high (7) \\
        Scale from weak (3) to strong (7) \\
        Scale from negative (0) to positive (6)
    } \\
    \midrule

    \makecell[l]{
        Wiki-Aggression \\
        Wiki-Attack \\
        Wiki-Toxicity
    }
    & \makecell[l]{HuBi/\\Wikipedia Detox}
    & \makecell[c]{7303 \\ 7303 \\ 7101}
    & \makecell[c]{488 \\ 488 \\ 490}
    & \makecell[c]{1.8 \\ 1.8 \\ 1.6}
    & \makecell[l]{
        Friendly/neutral (0), Aggressive (1) \\
        Not attack/harassment (0), Att./harass. (1) \\
        Healthy/neutral (0), Toxic (1)
    } \\
    \midrule

    \makecell[l]{SChem \\ SChem5Labels}
    & \makecell[l]{NLPerspectives}
    & \makecell[c]{400 \\ 7861}
    & \makecell[c]{100 \\ 102}
    & \makecell[c]{50.0 \\ 5.0}
    & \makecell[l]{
        No one believes (0), \\ 
        Occasionally believed (1), \\
        Controversial (2), Common belief (3), \\
        Universally true (4)
    } \\
    \midrule

    SBIC
    & NLPerspectives
    & 10000
    & 204
    & 2.5
    & \makecell[l]{
        Not offensive (0), Maybe (1), Offensive (2)
    } \\
    \midrule

    GHC
    & NLPerspectives
    & 10000
    & 18
    & 3.1
    & \makecell[l]{
        Not hate speech (0), Hate speech (1)
    } \\
    \midrule

    Sentiment
    & NLPerspectives
    & 8411
    & 500
    & 1.7
    & \makecell[l]{
        Very negative (0), Somewhat negative (1), \\
        Neutral (2), Somewhat positive (3), \\
        Very positive (4)
    } \\
    \midrule

    Politeness
    & \makecell[l]{
        NLPerspectives/\\
        Stanford Politeness\\
        Corpus
    }
    & 4338
    & 219
    & 5.0
    & \makecell[l]{
        Scale from very impolite (0)\\
        to very polite (24)
    } \\
    \midrule

    ConvAbuse
    & LeWiDi
    & 2398
    & 8
    & 3.0
    & \makecell[l]{
        Strongly negative with overt incitement \\
        to hatred, violence or discrimination; \\
        attitude oriented at attacking or \\
        demeaning the target (0), \\
        Negative and insulting/abusive, \\
        aggressive attitude (1), \\
        Negative and impolite, mildly offensive \\
        but still conversational (2), \\
        Ambiguous, unclear (3), \\
        Non-abusive (4)
    } \\
    \midrule

    HS-Brexit
    & LeWiDi
    & 784
    & 6
    & 6.0
    & \makecell[l]{
        Not hate speech (0), Hate speech (h.s.) (1)
    } \\
    \midrule

    MD-Agreement
    & LeWiDi
    & 6591
    & 500
    & 3.9
    & \makecell[l]{
        Offensive (0), Non-offensive (1)
    } \\
    \midrule

    CommitmentBank
    & \makecell[l]{AHLV/\\CommitmentBank}
    & 1200
    & 496
    & 9.6
    & \makecell[l]{
        Certainly false (0), Uncertain (3), \\
        Certainly true (6)
    } \\
    \midrule

    \makecell[l]{
        MHS-Sentiment \\
        (Other) $\times$ 7 \\
        MHS-Attack/Defense \\
        MHS-Hate Speech
    }
    & \makecell[l]{
        AHLV/\\
        Measuring Hate\\
        Speech
    }
    & \makecell[c]{1362 \\ 1362 \\ 1362 \\ 1362}
    & \makecell[c]{494 \\ 494 \\ 494 \\ 494}
    & \makecell[c]{1.6 \\ 1.6 \\ 1.6 \\ 1.6}
    & \makecell[l]{
        Scale from very neg. (0) to very pos. (4) \\
        Scale from low (0) to high (4) \\
        Scale from defense (0) to attack (4) \\
        Not h.s. (0), Unclear (1), H.s. (2)
    } \\
    \midrule

    \makecell[l]{
        (Other) $\times$ 2 \\
        AwA-Intent \\
        AwA-Racism
    }
    & \makecell[l]{
        AHLV/\\
        Annotators with\\
        Attitudes
    }
    & \makecell[c]{626 \\ 626 \\ 626}
    & \makecell[c]{184 \\ 184 \\ 184}
    & \makecell[c]{5.5 \\ 5.5 \\ 5.5}
    & \makecell[l]{
        Scale from not offensive (1) to very off. (5) \\
        Scale from unintentional (1) to intent. (5) \\
        Scale from not racist (1) to very racist (5)
    } \\
    \midrule

    DynaSent
    & \makecell[l]{AHLV/\\DynaSent}
    & 6678
    & 385
    & 1.5
    & \makecell[l]{
        Negative (0), Neutral (1), Positive (2)
    } \\
    \bottomrule
    \end{tabular}

    \caption{Information and statistics on the \NumDatasets{} datasets used in our experiment. The statistics include the number of unique text instances (N), the number of unique annotators (M), and the average number of ratings per unique text. Within each source section, ``(Other) $\times$ K'' represents K datasets with identical statistics and label meanings.}
    \label{tab:datasets}
\end{table*}

\begin{table*}[t]
    \centering
    \small
    \setlength{\aboverulesep}{0.1ex}
    \setlength{\belowrulesep}{0.2ex}
    \setlength{\tabcolsep}{2.5pt}
    \renewcommand{\arraystretch}{0.94}
    \begin{tabular}{lll}
    \toprule
    \textbf{Dataset Source} & \textbf{Dataset(s)} & \textbf{Terms} \\
    \midrule

    AffectiveText
    & \makecell[l]{ANGER \\ DISGUST \\ FEAR \\ JOY \\ SADNESS \\ SURPRISE \\ VALENCE}
    & \href{https://web.eecs.umich.edu/~mihalcea/affectivetext/}{Public Release} \\
    \midrule

    HuBi
    & Cockamamie Gobbledegook
    & \href{https://github.com/limorigu/Cockamamie-Gobbledegook}{Public Release} \\
    \midrule

    HuBi/Personal Bias
    & \makecell[l]{ANGER \\ ANTICIPATION \\ DISGUST \\ EMOT.\ AROUSAL \\ FEAR \\ JOY \\ SADNESS \\ SURPRISE \\ TRUST \\ VALENCE}
    & \href{https://github.com/CLARIN-PL/personal-bias}{Public Release} \\
    \midrule

    \makecell[l]{HuBi/Wikipedia Detox \\ \citep{wulczyn_wikipedia_2017-1,wulczyn_wikipedia_2017} \\ \citep{thain_wikipedia_2017}}
    & \makecell[l]{Wiki-Aggression \\ Wiki-Attack \\ Wiki-Toxicity}
    & \makecell[l]{
        \href{https://figshare.com/articles/dataset/Wikipedia_Talk_Labels_Aggression/4267550}{CC0 1.0} \\
        \href{https://figshare.com/articles/dataset/Wikipedia_Talk_Labels_Personal_Attacks/4054689}{CC0 1.0} \\
        \href{https://figshare.com/articles/dataset/Wikipedia_Talk_Labels_Toxicity/4563973}{CC0 1.0}}
    \\
    \midrule

    NLPerspectives
    & \makecell[l]{SChem \\ SChem5Labels \\ SBIC \\ GHC \\ Sentiment}
    & \makecell[l]{
        \href{https://github.com/mbforbes/social-chemistry-101}{CC BY-SA 4.0} \\
        \href{https://github.com/mbforbes/social-chemistry-101}{CC BY-SA 4.0} \\
        \href{https://huggingface.co/datasets/allenai/social_bias_frames}{CC BY 4.0} \\
        \href{https://osf.io/edua3/metadata}{CC BY 4.0} \\
        \href{https://doi.org/10.7910/DVN/F6EMTS}{CC0 1.0}}
    \\
    \midrule

    NLPerspectives/Stanford Politeness Corpus
    & Politeness
    & \href{https://convokit.cornell.edu/documentation/stack_politeness.html}{CC BY 4.0} \\
    \midrule

    LeWiDi
    & \makecell[l]{ConvAbuse \\ HS-Brexit \\ MD-Agreement}
    & \makecell[l]{
        \href{https://github.com/amandacurry/convabuse/blob/main/LICENSE}{CC BY 4.0} \\
        \href{https://github.com/Le-Wi-Di/le-wi-di.github.io}{Public Release} \\
        \href{https://github.com/dhfbk/annotators-agreement-dataset?tab=readme-ov-file}{Research Use Only}}
    \\
    \midrule

    AHLV/CommitmentBank
    & CommitmentBank
    & \href{https://github.com/mcdm/CommitmentBank}{Public Release}
    \\
    \midrule

    AHLV/Measuring Hate Speech
    & \makecell[l]{
        MHS-Sentiment, MHS-Respect \\
        MHS-Insult, MHS-Humiliation \\
        MHS-Status, MHS-Dehumanization \\
        MHS-Violence, MHS-Genocide \\
        MHS-Attack/Defense, MHS-Hate Speech}
    & \href{https://huggingface.co/datasets/ucberkeley-dlab/measuring-hate-speech}{CC BY 4.0}
    \\
    \midrule

    AHLV/Annotators with Attitudes
    & \makecell[l]{
        AwA-Offensive to You \\
        AwA-Offensive to Anyone \\
        AwA-Intent \\
        AwA-Racism}
    & \href{https://maartensap.com/racial-bias-hatespeech/annWithAttitudes-Qual.html}{Qualified Researchers}
    \\
    \midrule

    AHLV/DynaSent
    & DynaSent
    & \href{https://github.com/cgpotts/dynasent}{CC BY 4.0}
    \\
    \bottomrule
    \end{tabular}
    \caption{Publicly stated licenses, access restrictions, or availability terms for English-language dataset sources used in our experiments.}
    \label{tab:dataset-terms}
\end{table*}

A full enumeration of datasets used can be found in Table~\ref{tab:datasets}, while Table~\ref{tab:dataset-terms} lists the publicly stated licenses, access restrictions, or availability terms associated with the dataset sources used in our experiments.

\begin{figure*}[ht]
    \centering
    \includegraphics[width=0.49\textwidth]{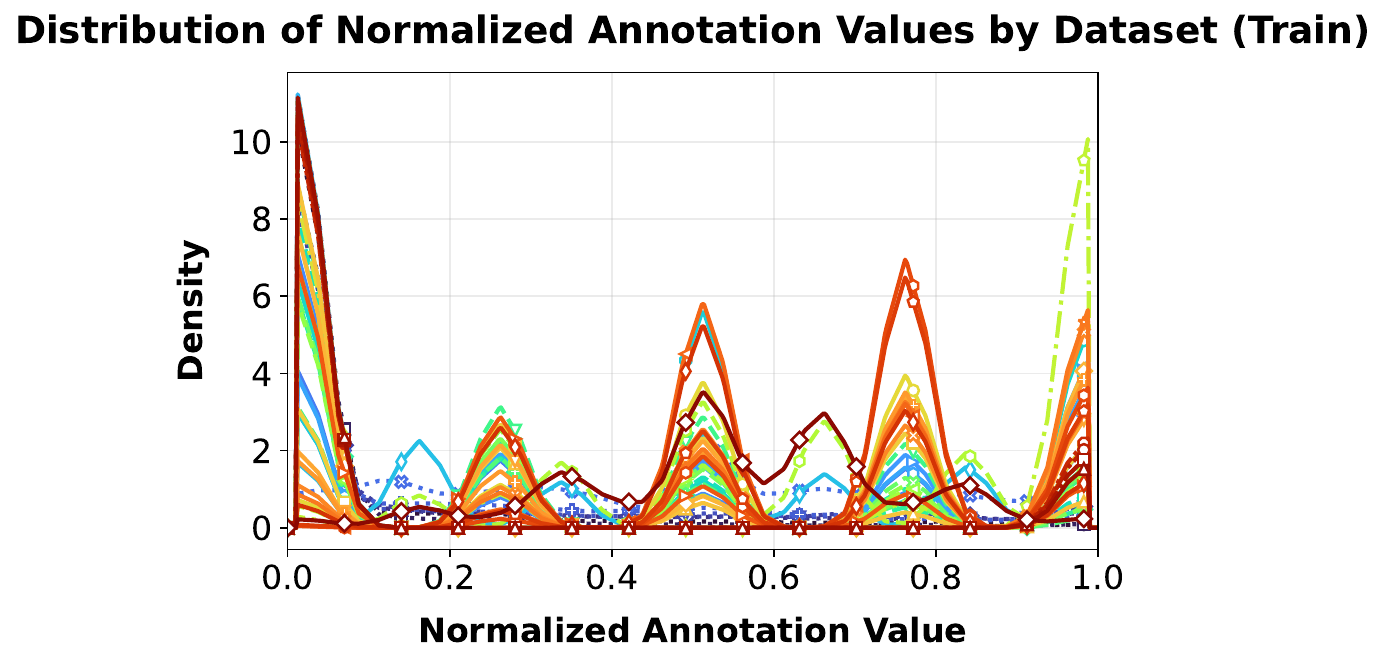}
    \includegraphics[width=0.49\textwidth]{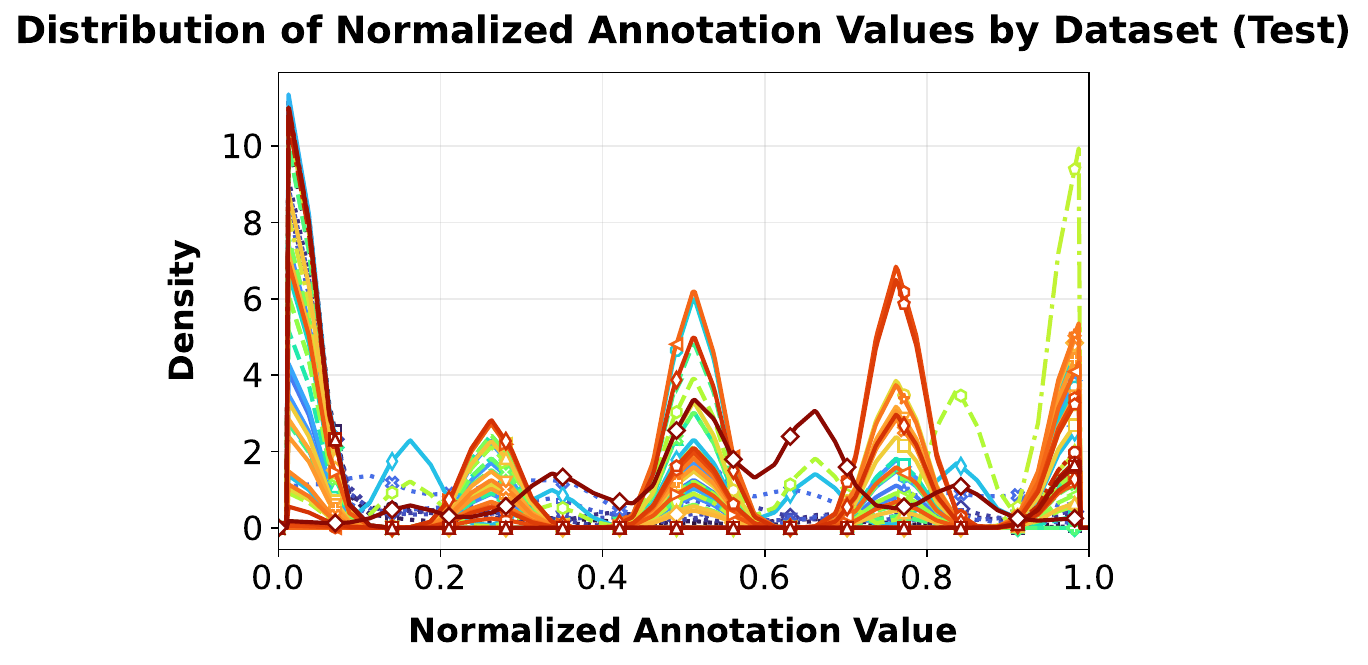}
    \includegraphics[width=\textwidth]{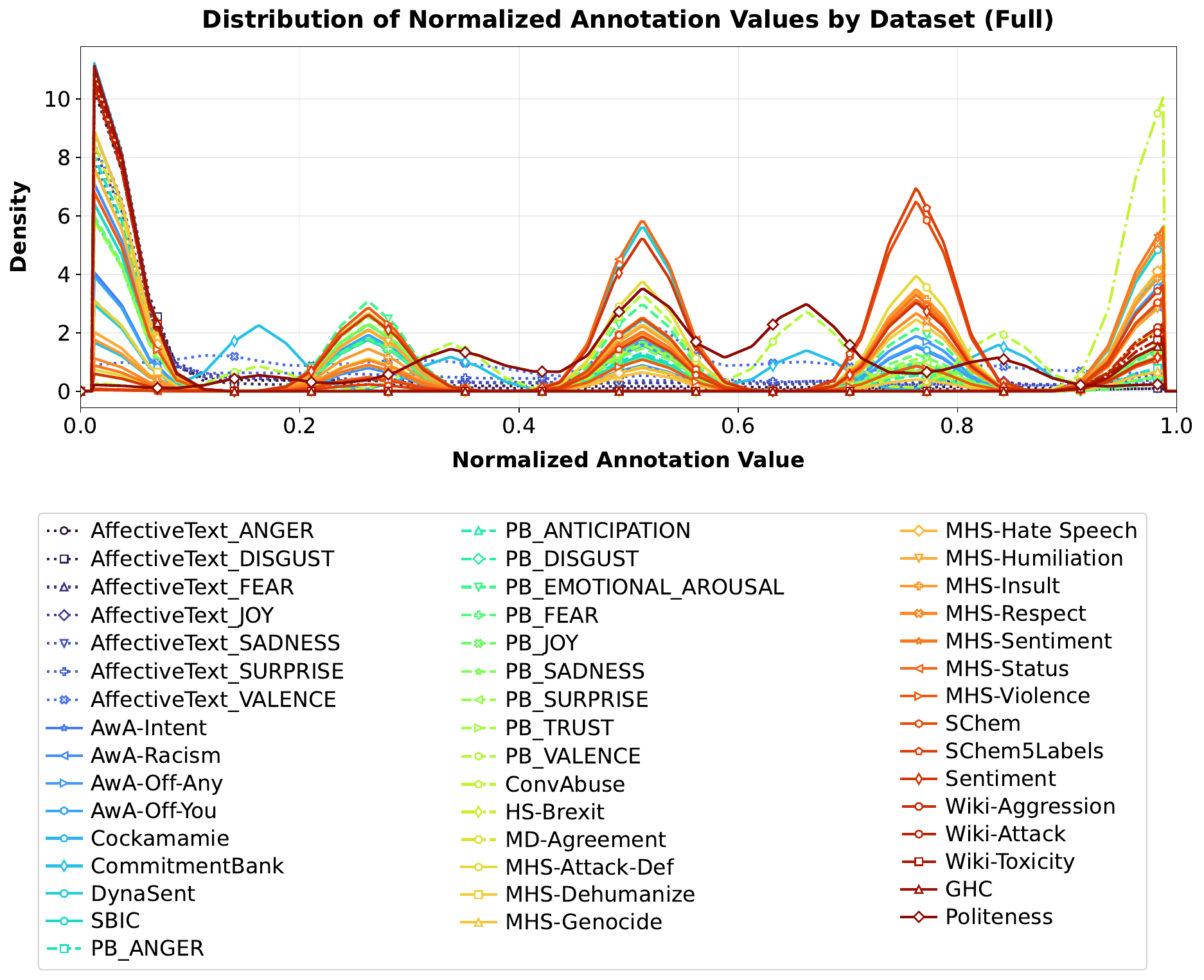}
    \caption{Smoothed density curves of annotation values across datasets with train data (top left), test data (top right), and both train and test data combined (bottom).}
    \label{fig:annotation-dist}
\end{figure*}

Figure~\ref{fig:annotation-dist} shows the distribution of annotation values across all datasets. We can observe large tails on each end of the distribution which are amplified by the presence of binary datasets.

\begin{figure*}[ht!]
    \centering
    \includegraphics[width=0.75\linewidth]{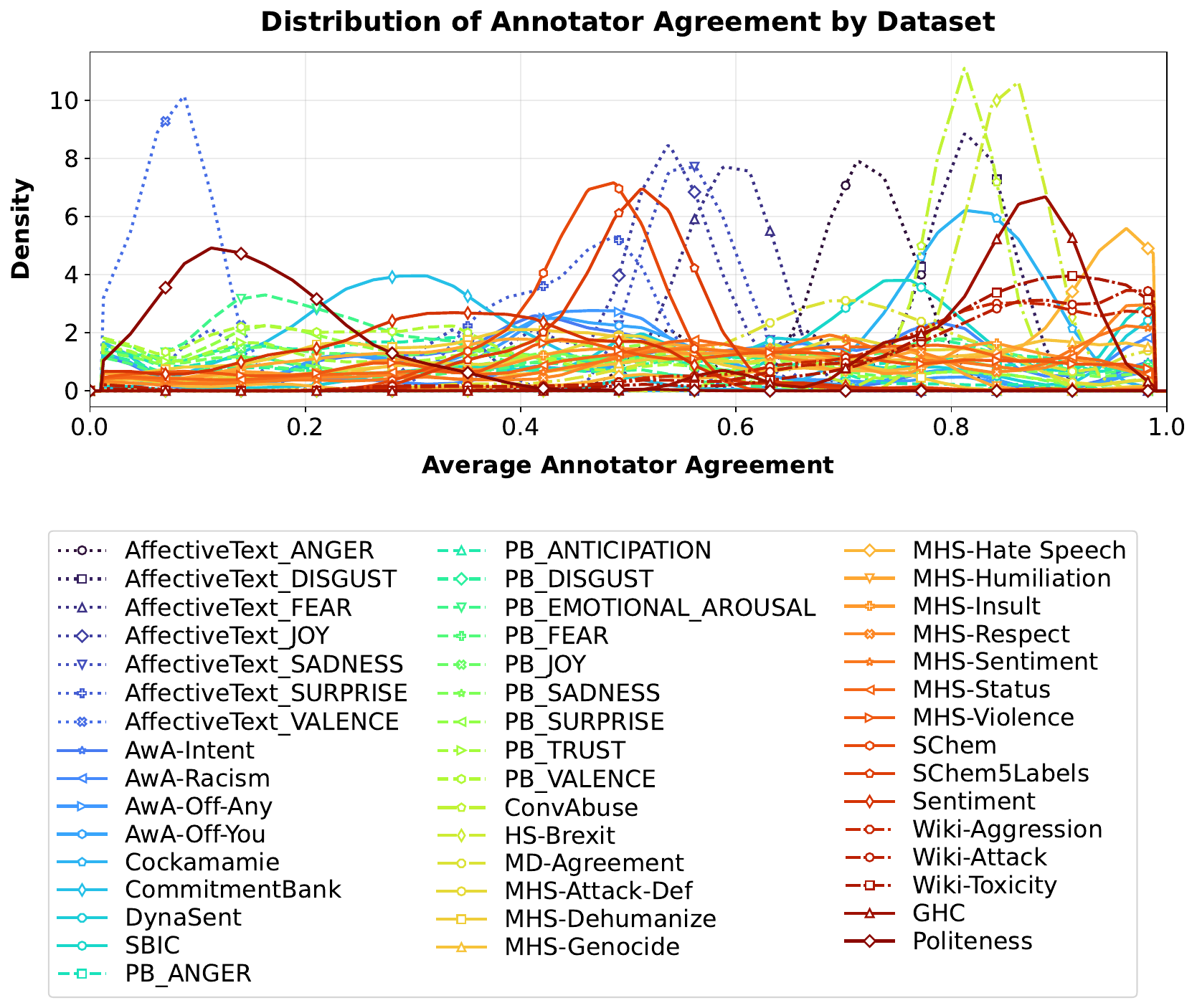}
    \includegraphics[width=0.75\linewidth]{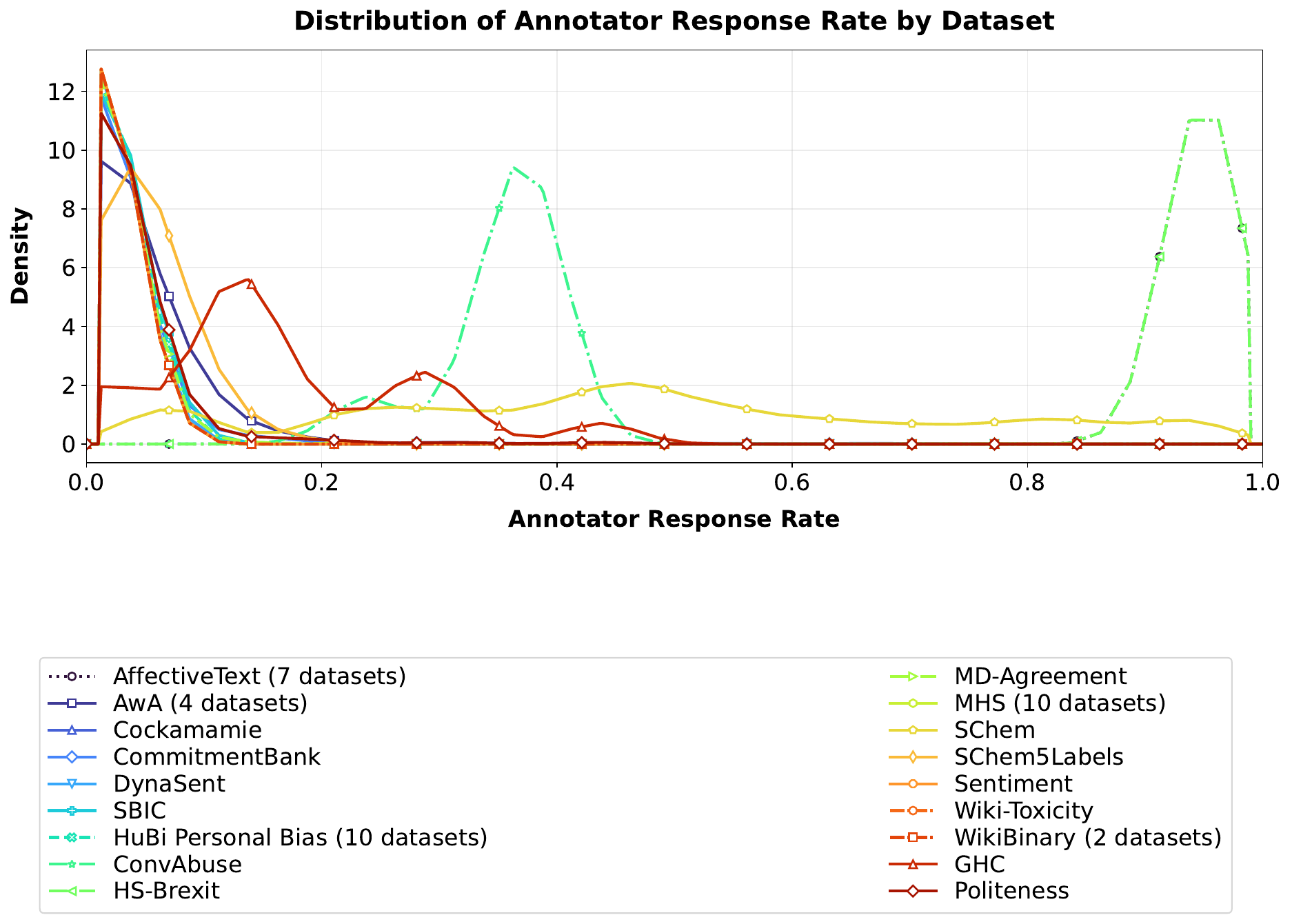}
    \caption{Distribution of agreement and annotator response rate across all datasets. Datasets with the same source and annotation patterns are grouped together when graphing response rate.}
    \label{fig:annotator-data-dist}
\end{figure*}

Figure~\ref{fig:annotator-data-dist} displays the distribution of agreement and annotator response rate for each dataset.

The smoothed density curves in the figure are computed over $[0, 1]$ by forming a 40-bin density-normalized histogram, smoothing it with a normalized discrete Gaussian-like kernel over integer offsets from -4 to 4 with scale 1.25, and interpolating the result onto a 400-point plotting grid. Near-constant distributions are visualized with a narrow normalized Gaussian fallback (bandwidth 0.015) centered at the observed value.

\subsection{SChem vs. SChem5Labels}
In this section, we clarify the relationship between the SChem and SChem5Labels datasets here, as we do not believe dataset provenance was covered adequately by the description given in \citet{lowmanstone_annotation_2023}.

The SChem and SChem5Labels datasets both come from \citet{forbes_social_2020}. SChem5Labels is a cleaned version of the ``dev-extra" and ``test-extra" splits of the released dataset merged together; each of these examples has 5 annotations per example. 

SChem consists of highly controversial examples (i.e. high annotator disagreement across the initial 5 annotators) from the ``analysis" split of the dataset, which were annotated with an additional 50 annotations per example. Thus, while SChem examples are a direct subset of SChem5Labels (before downsampling), unlike other datasets with the same source, the two SChem datasets do not share annotators. Due to the overlap in examples and rating guidelines, we group them into the same dataset family when computing dataset-family blocks for the bootstrapping analysis.

\section{Supplementary Post-Hoc Analysis}
\label{apx:post-hoc-analysis}

\subsection{Model Performance}
\label{apx:algo-performance}

\begin{figure*}[ht]
    \centering
    
    \includegraphics[width=0.49\textwidth]{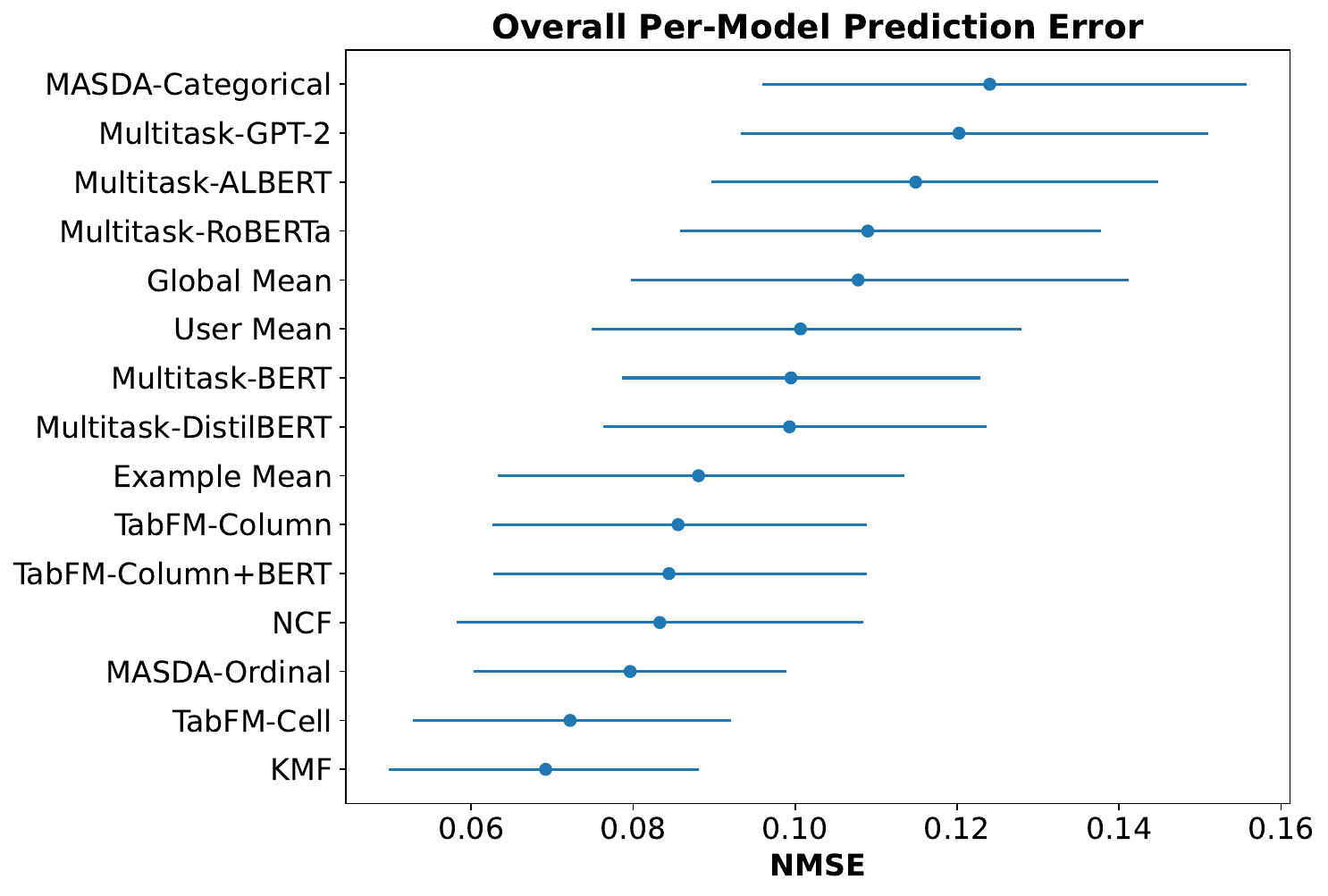}
    \includegraphics[width=0.49\textwidth]{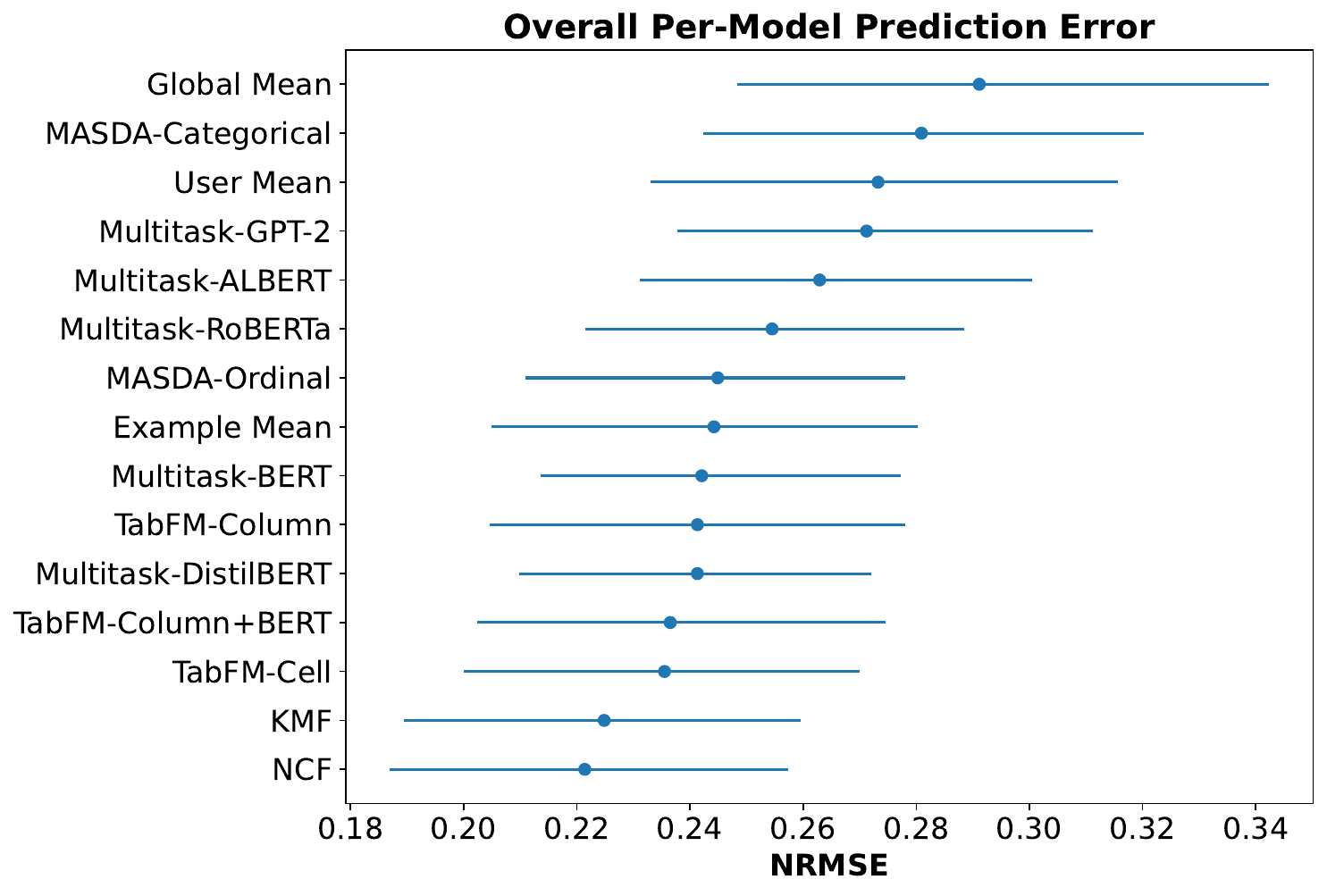}
    \includegraphics[width=0.75\textwidth]{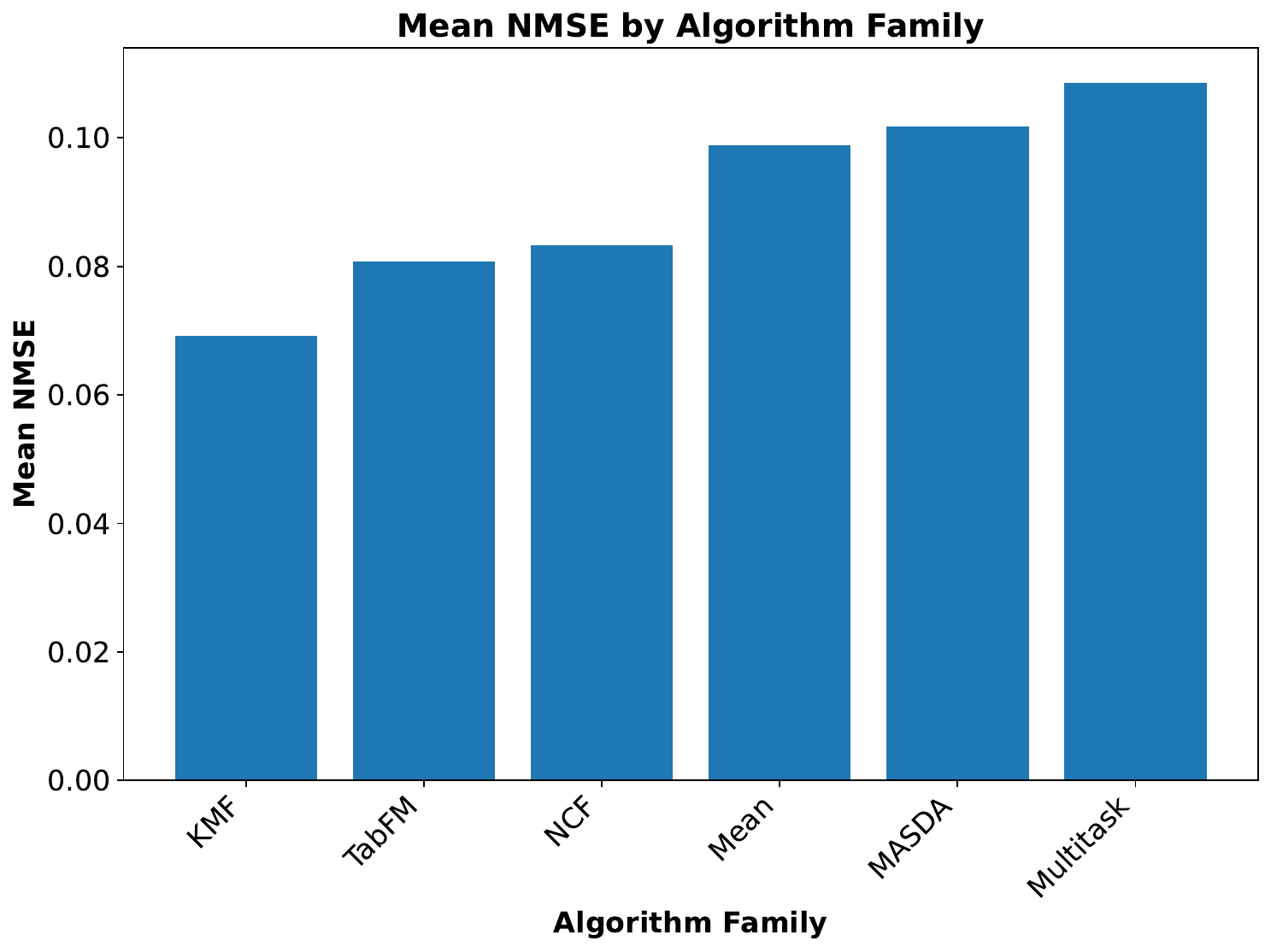}
    \caption{Forest plots of per-algorithm normalized mean squared error (top left), normalized root mean squared error (top right), and a bar chart of normalized mean squared error averaged over each algorithm family (bottom). The error bars in the forest plots show bootstrapped 95\% confidence intervals based on 1,000 resamples of individual datasets.}
    \label{fig:algo-performance}
\end{figure*}

\begin{figure*}[ht]
    \centering
    \includegraphics[width=0.95\textwidth]{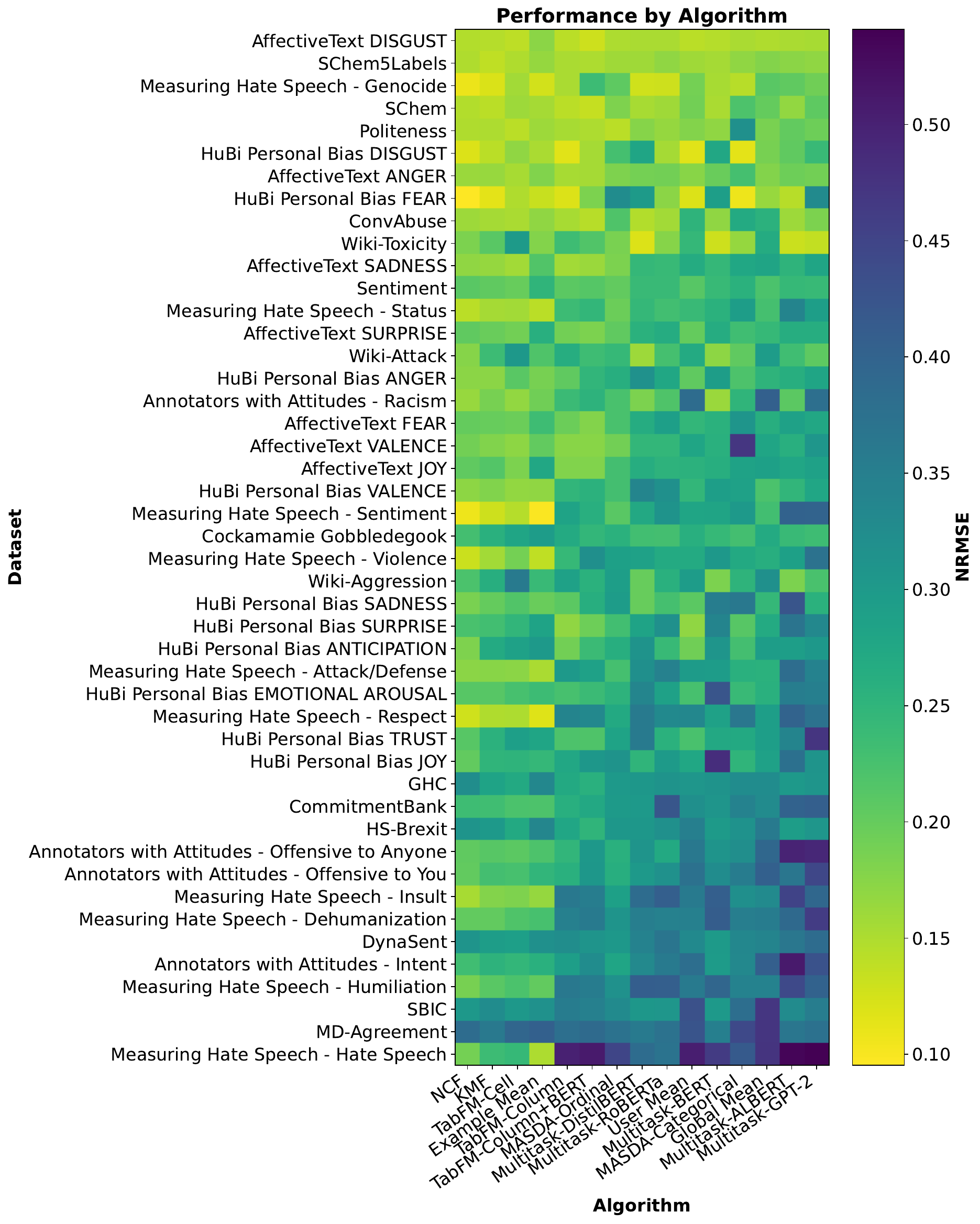}
    \caption{Per-model normalized root mean squared error performance on each dataset. Datasets are ordered from top to bottom in order of average performance across algorithms. Algorithms are ordered from left to right by average performance across datasets. Here, we use NRMSE (as opposed to NMSE) for better interpretability.}
    \label{fig:algo-dataset-heatmap}
\end{figure*}

\begin{figure*}[htbp]
    \includegraphics[width=0.95\textwidth]{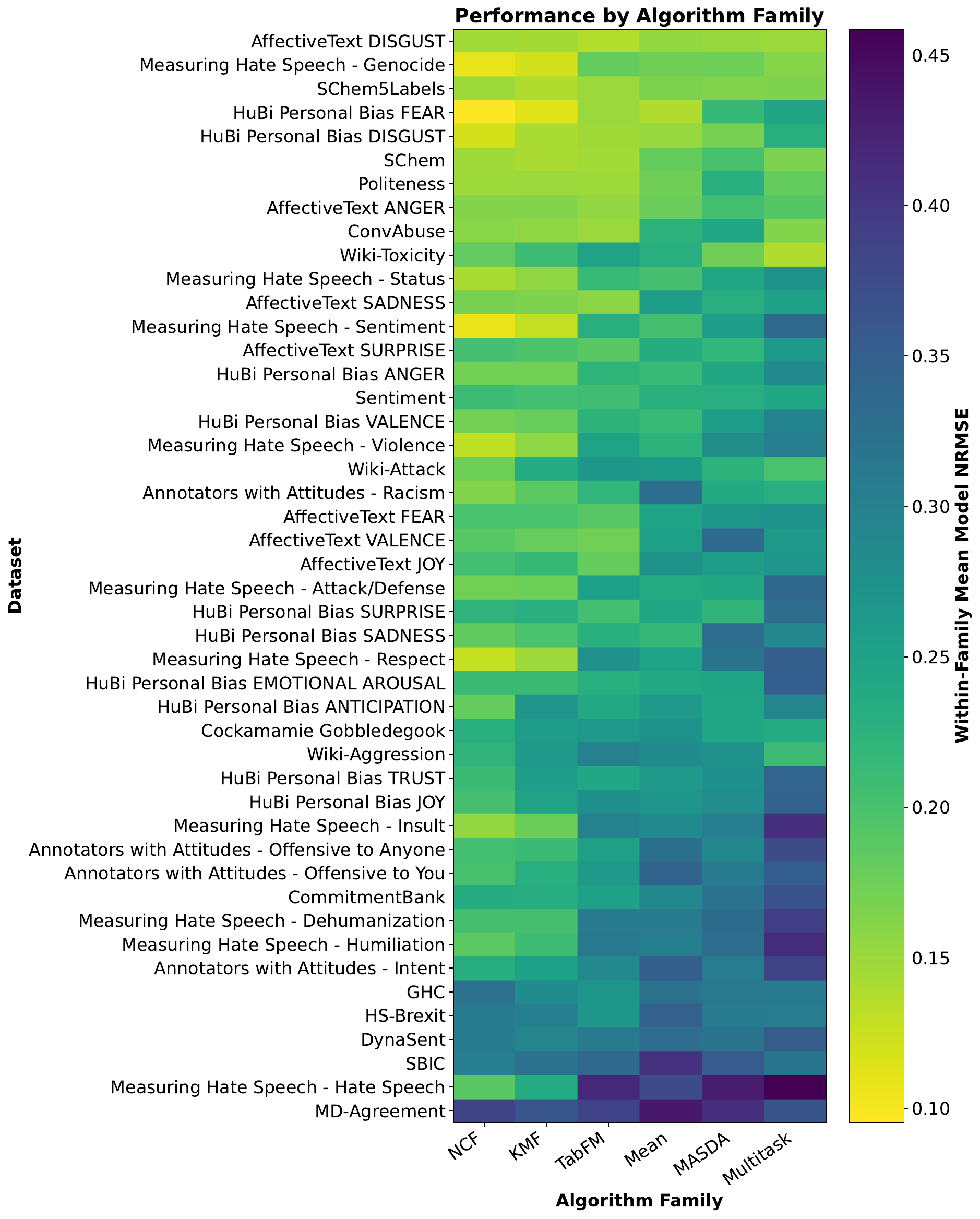}
    \caption{Average per-model-family normalized root mean squared error performance on each dataset. Similar to Figure~\ref{fig:algo-dataset-heatmap}, datasets are ordered from top to bottom by performance across algorithm families, algorithms are ordered from left to right by average performance across datasets, and we use NRMSE for stronger interpretability.}
    \label{fig:algo-family-dataset-heatmap}
\end{figure*}

Overall performance comparisons between algorithm families can be found in Figure~\ref{fig:algo-performance}. Heatmaps showing algorithm performance on individual datasets can be found in Figures~\ref{fig:algo-dataset-heatmap} and \ref{fig:algo-family-dataset-heatmap}.

\subsection{Two-Sided Test}
\label{apx:two-sided-test}

\begin{table*}[ht]
\centering
\begin{tabular}{lrrrr}
\toprule
Effect & Coef. & 95\% CI & Two-sided $p$ & Holm-adjusted $p$ \\
\midrule
Disagreement ($\beta_D$ on $Z^{D}_{da}$) &
\textbf{0.025} & \textbf{[0.016, 0.033]} &
$\mathbf{3.80 \times 10^{-8}}$ & $\mathbf{7.61 \times 10^{-8}}$ \\

Sparsity ($\beta_S$ on $Z^{S}_{da}$) &
0.001 & [-0.001, 0.003] &
0.317 & 0.317 \\
\bottomrule
\end{tabular}

\caption{Two-sided post hoc regression results for the disagreement and sparsity effects in the regression shown in Equation~\ref{eq:main-model} in the main paper. The main hypothesis tests shown in Table~\ref{tab:main-results} in the main paper use the predetermined one-sided tests for $\beta_D > 0$ and $\beta_S > 0$; these two-sided results are reported only to characterize the direction and robustness of the estimated effects. Results which are statistically significant are shown in \textbf{bold}.}
\label{tab:two-sided-test}
\end{table*}

The supplementary two-sided post hoc regression results for the disagreement and sparsity/response-rate effects are provided in Table~\ref{tab:two-sided-test}.

\end{document}